\documentclass[letterpaper]{article} % DO NOT CHANGE THIS
\usepackage{aaai24}  % DO NOT CHANGE THIS
\usepackage{times}  % DO NOT CHANGE THIS
\usepackage{helvet}  % DO NOT CHANGE THIS
\usepackage{courier}  % DO NOT CHANGE THIS
\usepackage[hyphens]{url}  % DO NOT CHANGE THIS
\usepackage{graphicx} % DO NOT CHANGE THIS
\usepackage{natbib}  % DO NOT CHANGE THIS AND DO NOT ADD ANY OPTIONS TO IT
\usepackage{caption} % DO NOT CHANGE THIS AND DO NOT ADD ANY OPTIONS TO IT
\usepackage{algorithm}
\usepackage{algorithmic}
\usepackage{bm}
\usepackage{newfloat}
\usepackage{listings}
\DeclareCaptionStyle{ruled}{labelfont=normalfont,labelsep=colon,strut=off} % DO NOT CHANGE THIS
\floatstyle{ruled}
\newfloat{listing}{tb}{lst}{}
\floatname{listing}{Listing}
\title{Painterly Image Harmonization by Learning from Painterly Objects}
\author{
    Li Niu\thanks{Corresponding author.},
    Junyan Cao, 
    Yan Hong, 
    Liqing Zhang 
    \\
}
\affiliations{
    MoE Key Lab of Artificial Intelligence, Shanghai Jiao Tong University\\
    \{ustcnewly, joy\_c1, hy2628982280, lqzhang\}@sjtu.edu.cn
}
\usepackage{bibentry}
\begin{document}

\maketitle

\begin{abstract}
Given a composite image with photographic object and painterly background, painterly image harmonization targets at stylizing the composite object to be compatible with the background. Despite the competitive performance of existing painterly harmonization works, they did not fully leverage the painterly objects in artistic paintings. In this work, we explore learning from painterly objects for painterly image harmonization. In particular, we learn a mapping from background style and object information to object style based on painterly objects in artistic paintings. With the learnt mapping, we can hallucinate the target style of composite object, which is used to harmonize encoder feature maps to produce the harmonized image. Extensive experiments on the benchmark dataset demonstrate the effectiveness of our proposed method. Dataset and code are available at \url{https://github.com/bcmi/ArtoPIH-Painterly-Image-Harmonization.} 
\end{abstract}

\section{Introduction}

Compositing multiple regions from different images is a fundamental photo editing technique. However, when pasting the foreground object extracted from one image on another background image, there may exist style inconsistency between foreground and background. To address such style inconsistency, myriads of image harmonization methods \cite{tsai2017deep,cong2020dovenet,ling2021region} have been developed to match the illumination information between foreground and background. 
Nevertheless, they are less effective when the foreground object is extracted from a photographic image and the background is an artistic painting, because complex styles (color, texture, pattern, strokes) need to be matched between foreground and background. 
Image harmonization with photographic object and painterly background is called painterly image harmonization \cite{luan2018deep}, which has only received limited attention. 

The existing painterly image harmonization methods can be roughly divided into optimization-based methods \cite{luan2018deep,zhang2020deep} and feed-forward methods \cite{peng2019element,cao2022painterly,yan2022style}. Optimization-based methods update the composite object iteratively to minimize the designed losses, which is too slow for real-time application. In contrast, feed-forward methods pass the composite image through the network only once to produce a harmonized image, which is much more efficient than optimization-based methods. To name a few,
\cite{peng2019element} introduced AdaIN \cite{huang2017arbitrary} to adjust the style of composite object.  \cite{cao2022painterly} explored harmonizing the composite image in both spatial domain and frequency domain. Some diffusion model based methods~\cite{sdedit,cdc} for cross-domain image composition can also be applied to this task. However, the harmonized objects may still look unnatural on the painterly background. 

\begin{figure}[t]
\centering
\includegraphics[width=1.0\linewidth]{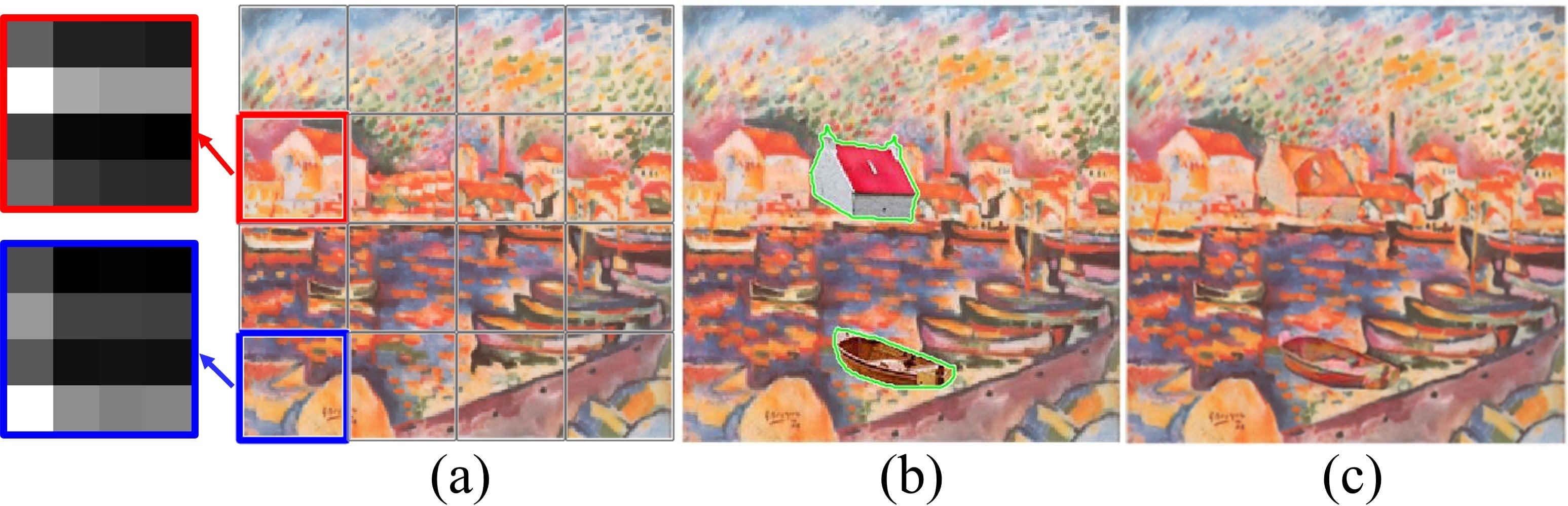}
\caption{(a) The background image with two marked query patches (red and blue). We show the similarity maps of two query patches based on VGG-19 \cite{VGG19} features. (b) The composite image with two inserted foreground objects. (c) The ideal harmonized image.}
\label{fig:illustration}
\end{figure}

One factor that hinders painterly image harmonization is the absence of ground-truth harmonized objects, but we can learn what an object in an artistic painting should be like based on the existent painterly objects in the artistic paintings. Therefore, this work explores learning from painterly objects, which is substantially overlooked by previous works \cite{luan2018deep,peng2019element,cao2022painterly,yan2022style}. The key insight of our work is hallucinating the target style of composite object. Previous works \cite{peng2019element,cao2022painterly} migrate the global style of painterly background to the composite object using AdaIN \cite{huang2017arbitrary}, during which the style is represented by the feature statistics (mean, variance) in pretrained VGG-19 \cite{VGG19} encoder. Nevertheless, the styles of different regions/objects in an artistic painting could vary considerably. As shown in Figure \ref{fig:illustration}(a), we divide the whole image into $16$ patches and extract the style vectors (mean/variance of VGG-19 encoder features) of all patches. We calculate their pairwise similarity based on style vectors. Given each query patch, we display a similarity map containing the similarities between this query patch and all patches. 
The patches with high similarity values have similar visual styles with the query patch. 
According to the similarity map, we can see that the style vector has prominent locality property, that is, the style vectors of various regions in the same image could be dramatically distinct. 
Therefore, when compositing different objects (\emph{e.g.}, house, boat) at different locations on the background, they should be harmonized to different target styles (see Figure \ref{fig:illustration}(b) and (c)). 

\emph{To achieve the goal of target style hallucination, we learn a mapping module based on painterly objects, which maps background style and object feature to object style.} The object features are supposed to contain the useful conditional information (\emph{e.g.}, color, semantics, local context) for adapting background style to object style. With the learnt mapping module, we attempt to hallucinate the target style of composite object based on its object feature and the background style. Intuitively, the mapping module answers the following question: ``\emph{given a painterly background, if its painter draws a specific object at this location, what should it look like?}"

However, one critical issue is the domain gap of object features between photographic objects and painterly objects, so that the mapping module trained on painterly objects cannot work well on composite objects. We have tried unsupervised manner like adversarial learning \cite{goodfellow2020generative} to bridge the domain gap, but the performance is far from satisfactory. Therefore, we choose to annotate pairs of similar photographic objects and painterly objects. Given an artistic painting with a painterly object, we retrieve and annotate photographic objects which are similar to this painterly object, leading to abundant triplets (artistic painting, painterly object, photographic object). In each triplet, we refer to the painterly object (\emph{resp.}, artistic painting) as the reference object (\emph{resp.}, reference image) of the photographic object. We put the photographic object within the bounding box of reference object in the reference image, resulting in a composite image (see Figure~\ref{fig:pair_construction}). 
In this way, we can acquire abundant pairs of composite images and reference images, in which the composite object feature should be close to reference object feature and the hallucinated style of composite object should be close to reference object style. With the above assumptions, we design a novel network with mapping module, and train it using pairs of composite images and reference images. 

Our major contributions can be summarized as follows. 1) We explore learning from painterly objects in the painterly image harmonization task, which has never been studied before. 2) We design a novel network with mapping module, which can hallucinate the target style of composite object based on the background style and object feature. 3) We will release our annotated reference images/objects, which would greatly benefit the future research of painterly image harmonization. 
4) Extensive experiments on COCO \cite{lin2014microsoft} and WikiArt \cite{nichol2016painter} demonstrate the effectiveness of our proposed method.

\section{Related Work}

\subsection{Image Harmonization} \label{sec:image_harmonization}

Image harmonization aims to harmonize a composite image by adjusting foreground illumination to match background illumination. 
In recent years, abundant deep image harmonization methods \cite{tsai2017deep,Jiang_2021_ICCV,xing2022composite,peng2022frih,zhu2022image,valanarasu2022interactive,LEMaRT} have been developed. For example, \cite{xiaodong2019improving,Hao2020bmcv,sofiiuk2021foreground} proposed diverse attention modules to treat the foreground and background separately, or establish the relation between foreground and background. \cite{cong2020dovenet,cong2021bargainnet,ling2021region,hang2022scs} directed image harmonization to domain translation or style transfer by treating different illumination conditions as different domains or styles.  \cite{guo2021image,guo2021intrinsic,guo2022transformer}  decomposed an image into reflectance map and illumination map.  More recently, \cite{cong2022high,ke2022harmonizer,liang2021spatial,xue2022dccf,PCTNet,WangCVPR2023} used deep network to predict color transformation, striking a good balance between efficiency and effectiveness. 

However, most image harmonization methods only adjust illumination and require ground-truth supervision, which is unsuitable for painterly image harmonization.

%-------------------------------------------------------------------------
\subsection{Painterly Image Harmonization}

Different from Section \ref{sec:image_harmonization}, in painterly image harmonization, the foreground is a photographic object while the background is artistic painting. The goal of painterly image harmonization is adjusting the foreground style to match background style and preserving the foreground content. The existing methods \cite{luan2018deep,zhang2020deep,peng2019element} can be roughly categorized into optimization-based methods \cite{luan2018deep,zhang2020deep} and feed-forward methods \cite{peng2019element}.  The optimization-based methods \cite{luan2018deep,zhang2020deep} iteratively optimize over the composite foreground to minimize the designed loss functions. The feed-forward methods \cite{peng2019element,yan2022style,cao2022painterly} send the composite image through the harmonization network once and generate the harmonized image. %Without iterative optimization during inference, the feed-forward methods are far more efficient than the optimization-based methods. 
Some diffusion model based methods~\cite{sdedit,cdc} for cross-domain image composition can also harmonize the composite image. 

Our proposed method belongs to feed-forward methods. Different from previous works \cite{peng2019element,yan2022style,cao2022painterly}, we explore learning from painterly objects to hallucinate the target styles of composite objects.

\subsection{Artistic Style Transfer}
Artistic style transfer~\cite{kolkin2019style,jing2020dynamic,chen2021dualast,chen2017stylebank,sanakoyeu2018style,wang2020collaborative,li2018learning,chen2016fast,sheng2018avatar,gu2018arbitrary,zhang2019multimodal,chen2022toward,huo2021manifold} targets at recomposing a content image in the style of a style image. Amounts of works~\cite{huang2017arbitrary,li2017universal} focus on global style transfer. 
%For example, AdaIN~\cite{huang2017arbitrary} applied mean and standard deviation to shift and re-scale the normalized content feature.
%WCT~\cite{li2017universal} adopted two transformation steps including whitening and coloring to achieve style transfer. A linear transformation according to content and style features was proposed in \cite{li2019learning}. 
Recently, some works \cite{park2019arbitrary,liu2021adaattn,deng2022stytr2} turn to fine-grained local style transfer by establishing local correspondences between content image and style image.  
For example, non-local attention mechanism was adopted in SANet~\cite{park2019arbitrary} to attentively transfer style features to content features. AdaAttN~\cite{liu2021adaattn} proposed  novel attentive normalization  by combining AdaIN~\cite{huang2017arbitrary} and SANet~\cite{park2019arbitrary}.  StyTr$^2$~\cite{deng2022stytr2} applied transformer architecture to capture patch-wise features to solve content leak issue. 

The motivation of local style transfer \cite{elad2017style,li2016combining,park2019arbitrary,huo2021manifold} is seeking for relevant regions in the style image for a given content image and transferring the corresponding local style. \emph{Nevertheless, the existence of relevant regions is questionable, and the methods may not find the relevant regions accurately (see Section~\ref{sec:cmp_with_baseline})}.

\section{Our Method}

\begin{figure*}[t]
\centering
\includegraphics[width=0.85\linewidth]{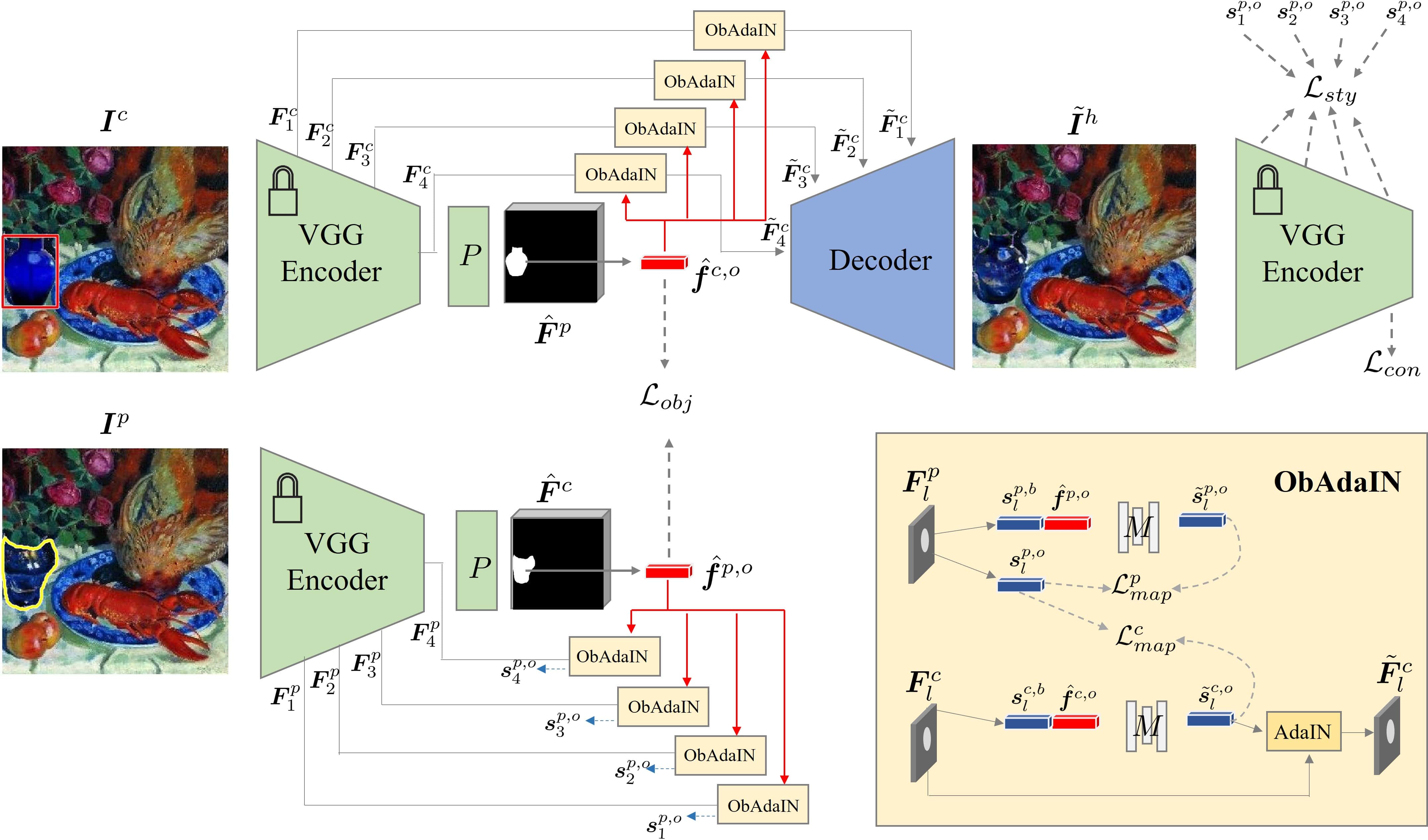}
\caption{The illustration of our network structure. Given a pair of composite image $\bm{I}^c$ and painterly image $\bm{I}^p$, we pass them through pretrained VGG-19~\cite{VGG19} encoder and projection module $P$ to get object features $\hat{\bm{f}}^{c,o}$ and $\hat{\bm{f}}^{p,o}$ respectively. We insert our designed ObAdaIN modules into skip connections and bottleneck, which harmonize the encoder feature maps of $\bm{I}^c$. The harmonized encoder feature maps are delivered to the decoder to produce the harmonized image $\tilde{\bm{I}}^h$. In the ObAdaIN module in the $l$-th layer, the mapping module $M_l$ learns a mapping from background style $\bm{s}_l^{p,b}$ (\emph{resp.}, $\bm{s}_l^{c,b}$) and object feature $\hat{\bm{f}}^{p,o}$ (\emph{resp.}, $\hat{\bm{f}}^{c,o}$) to object style $\tilde{\bm{s}}_l^{p,o}$  (\emph{resp.}, $\tilde{\bm{s}}_l^{c,o}$) for painterly (\emph{resp.}, composite) object. }
\label{fig:network}
\end{figure*}

We suppose that a composite image $\bm{I}^c$ is obtained by pasting a photographic object on the painterly background.  Painterly image harmonization aims to transfer the style from painterly background to composite object while preserving the content of composite object, resulting in a harmonized image $\tilde{\bm{I}}^h$. To learn from painterly objects in artistic paintings, we first build a training set with pairs of composite images $\bm{I}^c$ and reference images $\bm{I}^p$.
Then, we design a network which could be trained on paired training images. Our network is based on U-Net  \cite{ronneberger2015u} structure with skip connections, with our designed ObAdaIN module inserted into the bottleneck and skip connections. The ObAdaIN module aims to hallucinate the target style of composite object. 
We name our constructed training data as Arto (\textbf{Art}istic \textbf{O}bject) dataset and our method as ArtoPIH (\textbf{P}ainterly \textbf{I}mage \textbf{H}armonization). 
Next, we will introduce our Arto dataset in Section~\ref{sec:train_data_preparation} and our ArtoPIH in Section~\ref{sec:network_structure}. 

\subsection{Training Data Preparation}\label{sec:train_data_preparation}

\begin{figure}[t]
\centering
\includegraphics[width=0.9\linewidth]{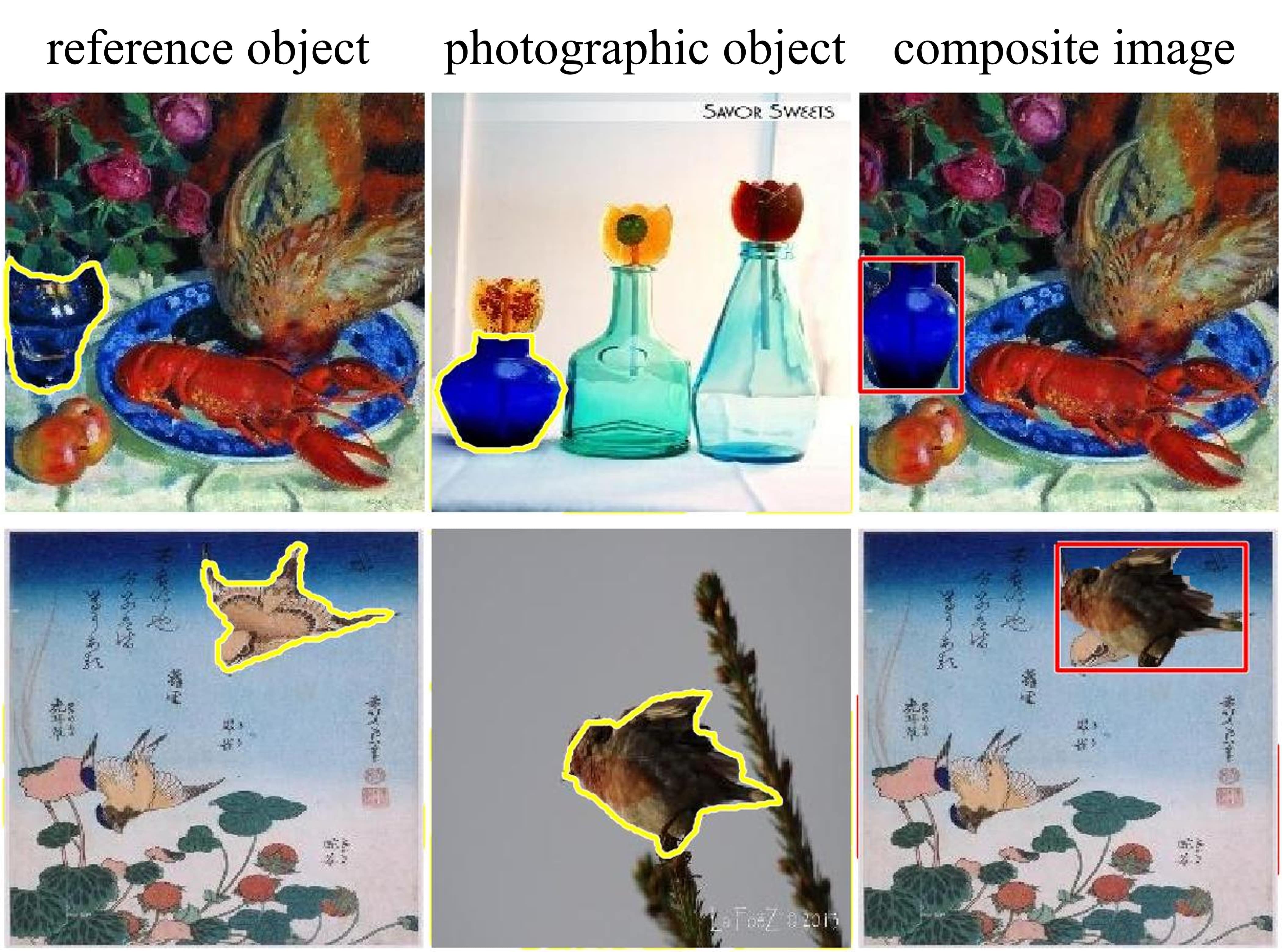}
\caption{The illustration of constructing training image pairs $\{\bm{I}^p, \bm{I}^c\}$. Given a photographic object (outlined in yellow), we have a reference object  (outlined in yellow) with similar color and semantics in the reference image $\bm{I}^p$. We put the photographic object within the bounding box of reference object, resulting in a composite image $\bm{I}^c$. }
\label{fig:pair_construction}
\end{figure}

Based on $57,025$ artistic paintings in the training set of WikiArt \cite{nichol2016painter}, we use off-the-shelf object detection model \cite{wu2019detectron2} pretrained on COCO \cite{lin2014microsoft} dataset to detect objects in artistic paintings. Despite the domain gap between artistic paintings and photographic images, we detect $34,570$ painterly objects. 
For each painterly object, we first retrieve $100$ photographic objects with similar appearance and semantics automatically (see details in the Supplementary). 

However, the retrieved photographic objects are very noisy, so we ask annotators to manually remove those dissimilar photographic objects. After filtering, we have $33,294$ painterly objects associated with similar photographic objects. 
%Among them, ``person" is the dominant category with $18,185$ painterly objects.
Each painterly object has an average of $9.83$ similar photographic objects, leading to a total of $327,181$  triplets (artistic painting $\bm{I}^p$, painterly object, photographic object). In each triplet, we refer to the painterly object (\emph{resp.}, artistic painting) as the reference object (\emph{resp.}, reference image) of the photographic object. As illustrated in Figure~\ref{fig:pair_construction}, we put the photographic object within the bounding box of reference object in the reference image, producing a composite image $\bm{I}^c$ with foreground mask $\bm{M}^c$. Therefore, we have in total $327,181$ training image pairs $\{\bm{I}^c, \bm{I}^p\}$. 

\subsection{Our Network Structure} \label{sec:network_structure}

We employ the encoder and decoder structures in \cite{huang2017arbitrary} as our backbone, in which the encoder is pretrained VGG-19~\cite{VGG19} and the decoder is symmetric structure of encoder. We only use the layers up to \emph{ReLU-4\_1} of VGG-19 as our encoder, which are fixed to extract multi-scale encoder features. The encoder is divided into four encoder layers, which are up to \emph{ReLU-1\_1}, \emph{ReLU-2\_1}, \emph{ReLU-3\_1}, and \emph{ReLU-4\_1} respectively. We add skip connections for the first three encoder layers. 

We feed composite image $\bm{I}^c$ into the encoder to extract the feature map $\bm{F}_l^c$ of $l$-th layer for \emph{l} $\in \left\{1, 2, 3, 4\right\}$. Similarly, we also feed its reference image $\bm{I}^p$ with reference object into the encoder to extract the feature map $\bm{F}^p_l$ of $l$-th layer for \emph{l} $\in \left\{1, 2, 3, 4\right\}$. 
Previous works on painterly image harmonization \cite{peng2019element,cao2022painterly} usually apply vanilla Adaptive Instance Normalization (AdaIN)~\cite{huang2017arbitrary} to the encoder feature maps, by aligning the feature statistics (mean, standard deviation) of foreground feature map with those of background feature map. As the feature statistics represent the style information, the composite object would be rendered with the background style. 

However, the background style may not be the ideal target style for the composite object, considering the diversified local styles within the same image and the unique property (\emph{e.g.}, color, semantics, local context) of composite object. In this work, we attempt to hallucinate the ideal target style of composite object by considering background style and object information. To achieve this goal, we design a novel Object-aware AdaIN (ObAdaIN) module to replace the vanilla AdaIN operation. 

\textbf{Domain-Invariant Object Feature:} Our ObAdaIN module requires the object feature of foreground object, which contains the necessary conditional information for adapting background style to object style. 

Given a training image pair $\{\bm{I}^c, \bm{I}^p\}$, the composite object in $\bm{I}^c$ is photographic object but the reference object in $\bm{I}^p$ is painterly object, so their object features $\hat{\bm{f}}^{c,o}$ and $\hat{\bm{f}}^{p,o}$ belong to two different domains. As shown in Figure~\ref{fig:network}, we first project feature maps $\bm{F}^p_4$ and $\bm{F}^c_4$ to a common domain through a projection module $P$, yielding $\hat{\bm{F}}^p$ and $\hat{\bm{F}}^c$ respectively. Then, we perform average pooling within the foreground region on  $\hat{\bm{F}}^p$ (\emph{resp.}, $\hat{\bm{F}}^c$) to get the object feature $\hat{\bm{f}}^{p,o}$ (\emph{resp.}, $\hat{\bm{f}}^{c,o}$). 

Since the composite object and its reference object have similar appearance and semantics, we push close $\hat{\bm{f}}^{p,o}$ and $\hat{\bm{f}}^{c,o}$ using the following loss:
\begin{eqnarray}\label{eqn:loss_obj} 
\mathcal{L}_{obj} = \|\hat{\bm{f}}^{p,o}-\hat{\bm{f}}^{c,o}\|^2.
\end{eqnarray}

By pulling close the object features of similar objects from two domains,  we enforce the extracted object features to be domain-invariant. 

\textbf{ObAdaIN Module: }With extracted domain-invariant object features, our ObAdaIN module learns a mapping from background style and object feature to object style. The architecture of ObAdaIN is shown in Figure~\ref{fig:network}. 

By taking the reference image $\bm{I}^p$ and the $l$-th encoder layer as an example, we have the feature map $\bm{F}^p_l$. We extract the background style vector $\bm{s}^{p,b}_l=[\bm{\mu}^{p,b}_l, \bm{\sigma}^{p,b}_l]$, in which $\bm{\mu}^{p,b}_l$ (\emph{resp.}, $\bm{\sigma}^{p,b}_l$) is channel-wise mean (\emph{resp.}, standard deviation) of background feature map. 

We concatenate the background style vector $\bm{s}^{p,b}_l$ with object feature $\hat{\bm{f}}^{p,o}$, which are sent to a mapping module $M_l$ to generate the object style vector $\tilde{\bm{s}}^{p,o}_l$.
The above process can be represented by $\tilde{\bm{s}}^{p,o}_l=M_l(\bm{s}^{p,b}_l, \hat{\bm{f}}^{p,o})$.
For the reference object in  $\bm{I}^p$, we can obtain its ground-truth style vector $\bm{s}^{p,o}_l$ by calculating the channel-wise mean and standard deviation of foreground feature map. We use $\bm{s}^{p,o}_l$ to supervise $\tilde{\bm{s}}^{p,o}_l$ as follows,
\begin{eqnarray}
\mathcal{L}_{map}^p = \|\tilde{\bm{s}}^{p,o}_l-\bm{s}^{p,o}_l\|^2. 
\end{eqnarray}

Compared with $\bm{I}^p$, its paired composite image $\bm{I}^c$ has similar foreground object, the same background, and the same foreground placement (bounding box). Therefore, we assume that the ideal target style of composite object should be close to reference object style $\bm{s}^{p,o}_l$. 
Specifically, given $\bm{I}^c$, we extract its background style vector $\bm{s}^{c,b}_l$ and object feature $\hat{\bm{f}}^{c,o}$ in a similar way to $\bm{I}^p$. Then, we can have $\tilde{\bm{s}}^{c,o}_l=M_l(\bm{s}^{c,b}_l, \hat{\bm{f}}^{c,o})$, which is supervised by $\bm{s}^{p,o}_l$:
\begin{eqnarray}
\mathcal{L}_{map}^c = \|\tilde{\bm{s}}^{c,o}_l-\bm{s}^{p,o}_l\|^2. 
\end{eqnarray}

Next, we use the hallucinated style vector $\tilde{\bm{s}}^{c,o}_l=[\tilde{\bm{\mu}}^{c,o}_l, \tilde{\bm{\sigma}}^{c,o}_l]$ to harmonize the composite image  $\bm{I}^c$. By denoting the foreground (\emph{resp.}, background) feature map in $\bm{F}_l^c$ as $\bm{F}^{c,o}_l$ (\emph{resp.}, $\bm{F}^{c,b}_l$), we apply AdaIN to adjust $\bm{F}^{c,o}_l$ as follows,
\begin{eqnarray}\label{eqn:adain} 
    \tilde{\bm{F}}^{c,o}_{l} = \tilde{\bm{\sigma}}^{c,o}_l\frac{\bm{F}^{c,o}_l-\mu(\bm{F}^{c,o}_l)}{\sigma(\bm{F}^{c,o}_l)} + \tilde{\bm{\mu}}^{c,o}_l,
\end{eqnarray}
in which $\mu(\cdot)$ (\emph{resp.}, $\sigma(\cdot)$) means calculating the mean (\emph{resp.}, standard deviation) of the specified feature map. The adjusted foreground feature map $\tilde{\bm{F}}^{c,o}_{l}$ is combined with the unchanged background feature map $\bm{F}^{c,b}_l$ to form the harmonized feature map $\tilde{\bm{F}}^{c}_{l}$.  

The harmonized feature maps of all encoder layers are sent to the decoder to produce the harmonized image $\bm{I}^h$. Similar to \cite{cao2022painterly}, we also adopt blending layer to combine the output image $\bm{I}^h$ and the composite image $\bm{I}^c$ with predicted soft blending mask, giving rise to the final harmonized image $\tilde{\bm{I}}^h$. 

For the harmonized image $\tilde{\bm{I}}^h$, we expect its object style to match the reference object style $\bm{s}^{p,o}_l$, and thus design the style loss accordingly:
\begin{eqnarray}\label{eqn:style_loss}
\mathcal{L}_{sty} =\!\!\!\!\!\!\!\!\!\!\!\!&&\sum_{l=1}^{4}\|\mu\left(\phi_{l}(\tilde{\bm{I}}^h)\circ \bm{M}^c\right)-\bm{\mu}^{p,o}_l \|^2 \nonumber\\
&&\!\!\!\!\!\!\!\!\!\!\!\!+ \sum_{l=1}^{4}\|\sigma\left(\phi_{l}(\tilde{\bm{I}}^h)\circ \bm{M}^c\right)-\bm{\sigma}^{p,o}_l \|^2,
\end{eqnarray}
where each $\phi_{l}(\cdot)$ denotes the $l$-th encoder layer in VGG-19~\cite{VGG19} encoder, $\circ$ means element-wise product. 

We use content loss \cite{gatys2016image} to ensure that the foreground content is maintained:
\begin{equation}\label{eqn:content_loss}
    \mathcal{L}_{con} =\left\|\phi_4(\tilde{\bm{I}}^h)-\phi_4(\bm{I}^c)\right\|^2,
\end{equation}
in which $\phi_{4}(\cdot)$ has been defined in Eqn.~\ref{eqn:style_loss}.

The total loss can be written as
\begin{eqnarray} \label{eqn:total_loss}
\mathcal{L}_{total} = \mathcal{L}_{obj} \!+\! \lambda (\mathcal{L}_{map}^p \!+\! \mathcal{L}_{map}^c) \!+\!  \mathcal{L}_{sty} \!+\!  \mathcal{L}_{con}, %\nonumber
\end{eqnarray}
in which the hyper-parameter $\lambda$ is empirically set as $10$.

\section{Experiments}

\begin{figure*}[t]
\centering
\includegraphics[width=1.0\linewidth]{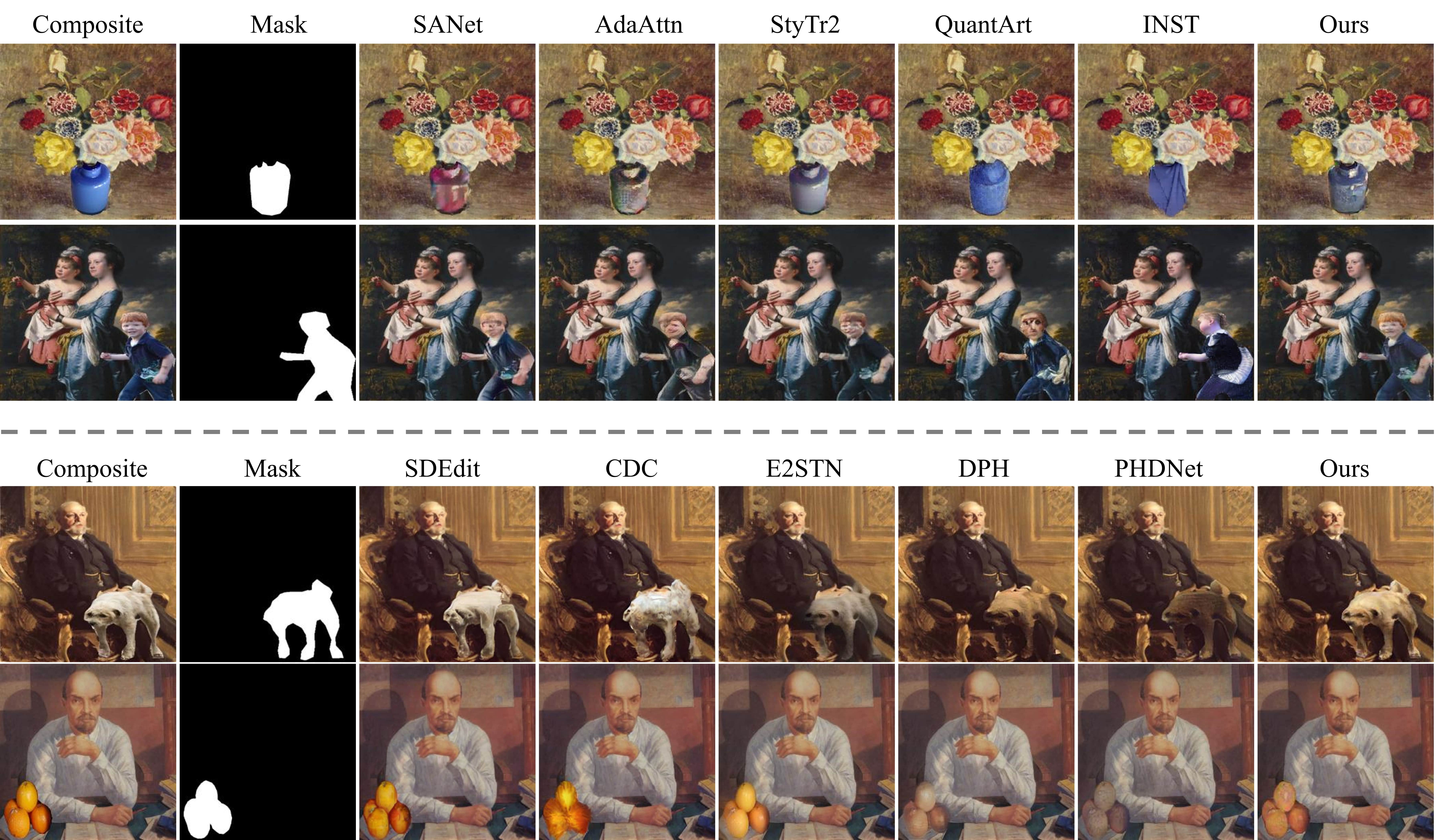}
\caption{In the upper part, we compare with style transfer baselines SANet~\cite{park2019arbitrary}, AdaAttN~\cite{liu2021adaattn}, StyTr2~\cite{deng2022stytr2}, QuantArt~\cite{quantart}, INST~\cite{inst}. 
In the lower part, we compare with painterly image harmonization baselines SDEdit~\cite{sdedit}, CDC~\cite{cdc}, E2STN~\cite{peng2019element}, DPH~\cite{luan2018deep}, PHDNet~\cite{cao2022painterly}.
 }
\label{fig:baseline_main}
\end{figure*}

\subsection{Datasets and Implementation Details} \label{sec:imp_detail}

We conduct experiments on COCO \cite{lin2014microsoft} and WikiArt \cite{nichol2016painter}. The details are left to Supplementary. 
Our network is implemented using Pytorch 1.10.0 and trained using Adam optimizer with learning rate of $1e{-4}$ on ubuntu 20.04 LTS operation system, with 128GB memory, Intel(R) Xeon(R) Silver 4116 CPU, and one GeForce RTX 3090 GPU. For the encoder and decoder structure, we follow \cite{cao2022painterly}. For $P$ module, we use one residual block \cite{he2016deep}. For $M_l$ module in the $l$-th encoder layer, we stack three ResMLP Layers \cite{TouvronBCCEGIJSVJ23}, in which the intermediate dimension is equal to the dimension of style vector in the $l$-th layer. 
%We set the batch size as $4$ and resize the input images as $256 \times 256$ during the training phase. 
%As our network is fully convolutional, it can be applied to images of any size in the test phase. Total of 100 epochs are used for convergence.

\subsection{Comparison with Baselines} \label{sec:cmp_with_baseline}

We compare with two groups of baselines: artistic style transfer methods and painterly image harmonization methods.  Since the standard image harmonization methods introduced in Section~\ref{sec:image_harmonization} only adjust the illumination statistics and require ground-truth images as supervision, they are not suitable for painterly image harmonization. 

For the first group of baselines, we first use artistic style transfer methods to stylize the entire photographic image according to the painterly background, and then paste the stylized photographic object on the painterly background.
We compare with typical or recent style transfer methods: SANet~\cite{park2019arbitrary}, AdaAttN~\cite{liu2021adaattn},  StyTr2~\cite{deng2022stytr2}, QuantArt~\cite{quantart}, INST~\cite{inst}. 
For the second group of baselines, we compare with SDEdit~\cite{sdedit}, CDC~\cite{cdc}, E2STN~\cite{peng2019element}, DPH~\cite{luan2018deep}, PHDNet~\cite{cao2022painterly}. 

\textbf{Visualization Results:} The comparison with the first group of baselines is shown in the upper part in Figure~\ref{fig:baseline_main}. We can see that the harmonized objects generated by our method look more natural and compatible with the background. In row 1, one interesting observation is that the vases harmonized by different methods have different colors. One possible reason is that SANet, AdaAttN, and StyTr2 seek for the relevance between composite object and different regions in the painterly background, and transfer the local style of relevant region to the composite object. However, there may not exist relevant regions in the painterly background. Even if relevant regions exist, they may not accurately attend to the relevant regions. In contrast, we hallucinate the target style based on background style and object information. Our harmonized vase maintains the object color and also has background texture, while the baseline results are over-smooth without compatible textures or losing some structure details (\emph{e.g.}, vase neck). In row 2, the harmonized faces from baseline methods all have corrupted structures, while our method preserves the facial structure well. This might be because that our model learns from abundant painterly person in artistic paintings and the hallucinated target style is able to preserve the facial structure. 

\begin{table}[t] 
\centering
\begin{tabular}{c|c|c|c}
\hline
Method  & B-T score & Time(s)  & FLOPs(G) \\
\hline
SANet  & -0.365 & 0.0097 & 43.32\\
AdaAttN  & -0.535 & 0.0115 & 49.64\\
StyTr2  &  0.149 & 0.0504 & 39.74\\
QuantArt   & 0.336 & 0.1031 & 133.34 \\
Inst  & -0.762 & 2.2996 & 3378.43 \\
\hline
SDEdit  & -0.654 & 2.1321 & 3164.52\\
CDC   &  0.204 & 2.3427 & 3299.81 \\
E2STN & -0.152 & 0.0079 & 29.28\\
DPH   & 0.340 & 55.24 & -\\
PHDNet  & 0.485 & 0.0321 & 158.41\\
\hline
ArtoPIH  & 0.953 & 0.0258 & 40.03\\
\hline
\end{tabular}
\caption{The comparison between different methods. }
\label{tab:results}
\end{table}

The comparison with the second group of baselines is shown in the lower part in Figure~\ref{fig:baseline_main}. In row 1, our harmonized dog has more harmonious color and compatible texture with the background, while the baseline results are corrupted or overdark. In row 2, our harmonized fruits have more suitable color and textures, and thus look more visually pleasant than the baseline results. More visualization results can be found in Supplementary.

\begin{figure*}[t]
\centering
\includegraphics[width=0.95\linewidth]{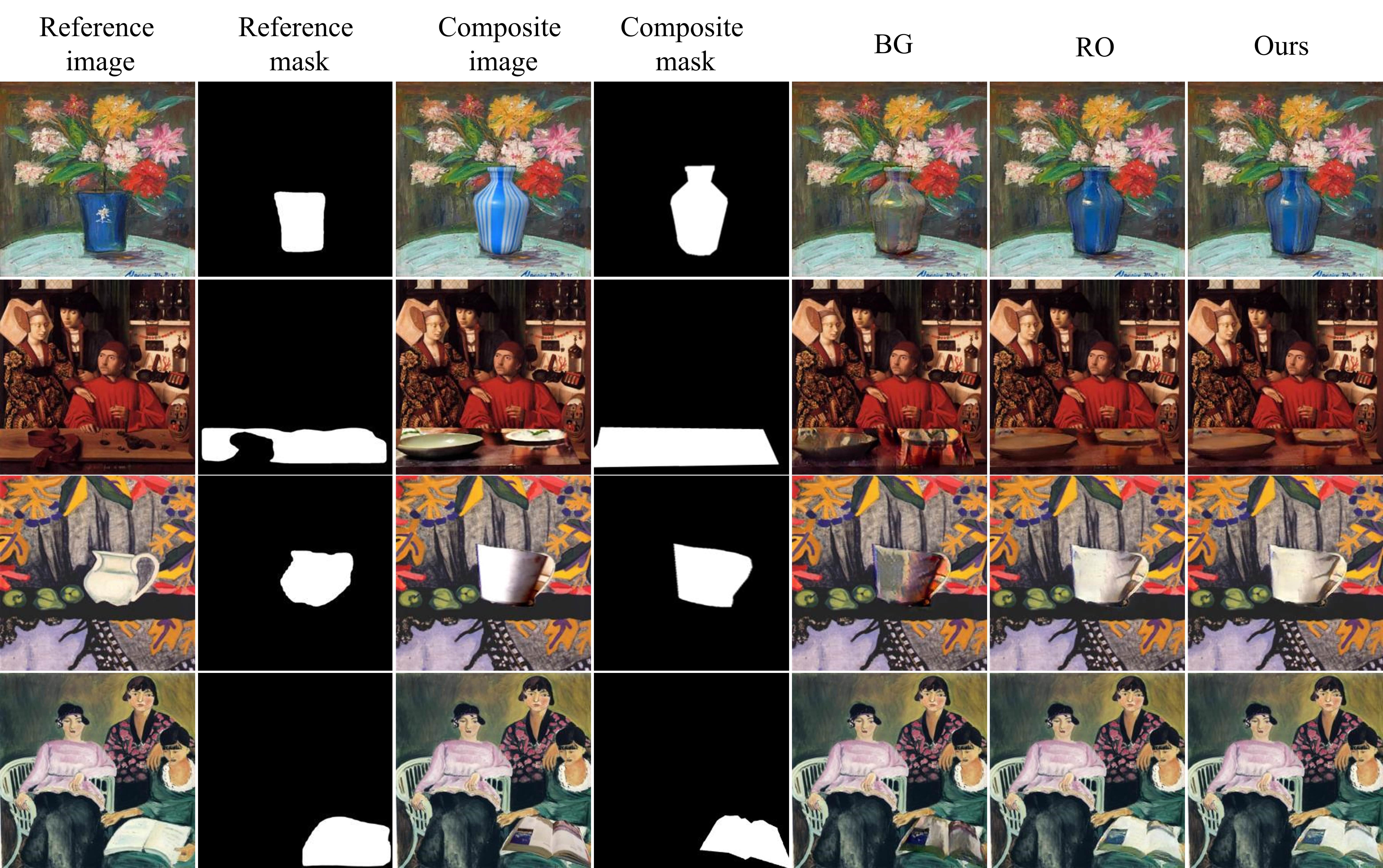}
\caption{From left to right, we show the reference image, the mask of reference object, the composite image, the mask of composite object, and the harmonized results obtained using different style vectors. ``BG" uses background style vector, ``RO" uses reference object style vector, and ``Ours" uses our hallucinated style vector.}
\label{fig:reference_results}
\end{figure*}

\textbf{User Study:} Following \cite{cao2022painterly}, we also conduct user study to compare different methods. We randomly select 100 content images from COCO and 100 style images from WikiArt to generate 100 composite images. We compare the harmonized results generated by 10 baselines and our ArtoPIH.

Specifically, for each composite image, we obtain $11$ harmonized outputs and use every two images from these $11$ images to construct image pairs. With $100$ composite images, we construct $5500$ image pairs. Then, we invite $100$ participants to watch one image pair each time and choose the better one. Finally, we collect $550,000$ pairwise results and calculate the overall ranking of all methods using Bradley-Terry (B-T) model~\cite{bradley1952rank,lai2016comparative}. The B-T score results are summarized in Table~\ref{tab:results}, in which our method achieves the highest B-T score.

\textbf{Efficiency Analyses:} For efficiency comparison, we report the inference time and FLOPs. We evaluate with the image size $256 \times 256$ and the inference time is averaged over 100 test images on a single RTX 3090 GPU.
The optimization-based method DPH is time-consuming, due to iteratively updating the input composite image. Diffusion-based methods are much slower than the other feed-forward methods, which limits their real-world applicability. 
Our method is relatively efficient compared with the competitive baselines \cite{cao2022painterly,luan2018deep}.

\subsection{Hallucinated Object Style}
To intuitively demonstrate the effectiveness of our hallucinated object styles, we sample some photographic objects in the test set and find their reference painterly objects in the test set, similar to the process of constructing training image pairs (see Section \ref{sec:train_data_preparation}). Specifically, given a photographic object and a painterly object in a painterly image, we can obtain a composite image as illustrated in Figure~\ref{fig:pair_construction}, followed by calculating the $L_2$ distance between the composite object feature $\hat{\bm{f}}^{c,o}$ and painterly object feature $\hat{\bm{f}}^{p,o}$. For each photographic object, we retrieve its nearest painterly object as reference object and the corresponding painterly image as reference image. In Figure \ref{fig:reference_results}, we show several examples of photographic objects and the retrieved reference objects. It can be seen that the photographic objects have similar color and semantics with their reference objects, which verifies the effectiveness of our learnt domain-invariant object features. %More examples of retrieved reference objects are provided in the Supplementary. 

Based on the composite image composed by the photographic object and its reference image, we further adjust the shape of photographic object to fully cover the reference object in the reference image, to exclude the influence of reference object. Then, based on our trained model, we apply three types of style vectors to the composite object to get the harmonized images.  
The first one is the background style vector $\bm{s}_l^{c,b}$. The second one is the style vector $\bm{s}_l^{p,o}$ of reference object. The third one is our hallucinated style vector $\tilde{\bm{s}}_l^{c,o}$. We apply the above three style vectors to the foreground feature map using AdaIN operation. The obtained harmonized images are denoted as ``BG", ``RO", and ``Ours" respectively, as shown in Figure~\ref{fig:reference_results}.
The harmonized results ``BG" and ``RO" differ a lot, which demonstrates the huge gap between background style and object style. Directly applying background style may destroy the original color of composite object or bring in noticeable artifacts. 

Since the composite object and reference object have similar object information and background style, the target style $\tilde{\bm{s}}_l^{c,o}$ of composite object is expected to approach the reference object style $\bm{s}_l^{p,o}$. Based on Figure~\ref{fig:reference_results},  the harmonized results ``Ours" are close to ``RO", which proves that our model can hallucinate ideal target styles. 

\subsection{More Results in Supplementary}

In the supplementary, we will provide ablation study results, more visual comparison with baselines, and discussion on failure cases. We will also show the results beyond COCO \cite{lin2014microsoft} dataset to demonstrate the generalization ability of our method across different datasets and different object categories. 

\section{Conclusion}
In this work, we have explored learning from painterly objects for painterly image harmonization. Based on the annotated pairs of composite images and reference painterly images, we have succeeded in hallucinating the target style of composite object, leading to visually pleasing harmonization results. Extensive experiments on the benchmark dataset have proved the advantage of our proposed ArtoPIH.

\section*{Acknowledgments}
The work was supported by the National Natural Science Foundation of China (Grant No. 62076162), the Shanghai Municipal Science and Technology Major/Key Project, China (Grant No. 2021SHZDZX0102, Grant No. 20511100300).

\bibliography{main.bbl}

\begin{thebibliography}{63}
\providecommand{\natexlab}[1]{#1}

\bibitem[{Bradley and Terry(1952)}]{bradley1952rank}
Bradley, R.~A.; and Terry, M.~E. 1952.
\newblock Rank analysis of incomplete block designs: I. The method of paired
  comparisons.
\newblock \emph{Biometrika}, 39(3/4): 324--345.

\bibitem[{Cao, Hong, and Niu(2023)}]{cao2022painterly}
Cao, J.; Hong, Y.; and Niu, L. 2023.
\newblock Painterly Image Harmonization in Dual Domains.
\newblock \emph{AAAI}.

\bibitem[{Chen et~al.(2017)Chen, Yuan, Liao, Yu, and Hua}]{chen2017stylebank}
Chen, D.; Yuan, L.; Liao, J.; Yu, N.; and Hua, G. 2017.
\newblock Stylebank: An explicit representation for neural image style
  transfer.
\newblock In \emph{CVPR}.

\bibitem[{Chen et~al.(2021)Chen, Zhao, Wang, Zhang, Zuo, Li, Xing, and
  Lu}]{chen2021dualast}
Chen, H.; Zhao, L.; Wang, Z.; Zhang, H.; Zuo, Z.; Li, A.; Xing, W.; and Lu, D.
  2021.
\newblock Dualast: Dual style-learning networks for artistic style transfer.
\newblock In \emph{CVPR}.

\bibitem[{Chen and Schmidt(2016)}]{chen2016fast}
Chen, T.~Q.; and Schmidt, M. 2016.
\newblock Fast patch-based style transfer of arbitrary style.
\newblock \emph{arXiv preprint arXiv:1612.04337}.

\bibitem[{Chen et~al.(2022)Chen, Wang, Xie, Lu, and Luo}]{chen2022toward}
Chen, Z.; Wang, W.; Xie, E.; Lu, T.; and Luo, P. 2022.
\newblock Towards Ultra-Resolution Neural Style Transfer via Thumbnail Instance
  Normalization.
\newblock In \emph{AAAI}.

\bibitem[{Cong et~al.(2021)Cong, Niu, Zhang, Liang, and
  Zhang}]{cong2021bargainnet}
Cong, W.; Niu, L.; Zhang, J.; Liang, J.; and Zhang, L. 2021.
\newblock BargainNet: Background-guided domain translation for image
  harmonization.
\newblock In \emph{ICME}.

\bibitem[{Cong et~al.(2022)Cong, Tao, Niu, Liang, Gao, Sun, and
  Zhang}]{cong2022high}
Cong, W.; Tao, X.; Niu, L.; Liang, J.; Gao, X.; Sun, Q.; and Zhang, L. 2022.
\newblock High-Resolution Image Harmonization via Collaborative Dual
  Transformations.
\newblock In \emph{CVPR}.

\bibitem[{Cong et~al.(2020)Cong, Zhang, Niu, Liu, Ling, Li, and
  Zhang}]{cong2020dovenet}
Cong, W.; Zhang, J.; Niu, L.; Liu, L.; Ling, Z.; Li, W.; and Zhang, L. 2020.
\newblock Dovenet: Deep image harmonization via domain verification.
\newblock In \emph{CVPR}.

\bibitem[{Cun and Pun(2020)}]{xiaodong2019improving}
Cun, X.; and Pun, C. 2020.
\newblock Improving the Harmony of the Composite Image by Spatial-Separated
  Attention Module.
\newblock \emph{{IEEE} Trans. Image Process.}, 29: 4759--4771.

\bibitem[{Deng et~al.(2022)Deng, Tang, Dong, Ma, Pan, Wang, and
  Xu}]{deng2022stytr2}
Deng, Y.; Tang, F.; Dong, W.; Ma, C.; Pan, X.; Wang, L.; and Xu, C. 2022.
\newblock StyTr2: Image Style Transfer with Transformers.
\newblock In \emph{CVPR}.

\bibitem[{Elad and Milanfar(2017)}]{elad2017style}
Elad, M.; and Milanfar, P. 2017.
\newblock Style transfer via texture synthesis.
\newblock \emph{IEEE Transactions on Image Processing}, 26(5): 2338--2351.

\bibitem[{Gatys, Ecker, and Bethge(2016)}]{gatys2016image}
Gatys, L.~A.; Ecker, A.~S.; and Bethge, M. 2016.
\newblock Image style transfer using convolutional neural networks.
\newblock In \emph{CVPR}.

\bibitem[{Goodfellow et~al.(2020)Goodfellow, Pouget-Abadie, Mirza, Xu,
  Warde-Farley, Ozair, Courville, and Bengio}]{goodfellow2020generative}
Goodfellow, I.; Pouget-Abadie, J.; Mirza, M.; Xu, B.; Warde-Farley, D.; Ozair,
  S.; Courville, A.; and Bengio, Y. 2020.
\newblock Generative adversarial networks.
\newblock \emph{Communications of the ACM}, 63(11): 139--144.

\bibitem[{Gu et~al.(2018)Gu, Chen, Liao, and Yuan}]{gu2018arbitrary}
Gu, S.; Chen, C.; Liao, J.; and Yuan, L. 2018.
\newblock Arbitrary style transfer with deep feature reshuffle.
\newblock In \emph{CVPR}.

\bibitem[{Guerreiro, Nakazawa, and Stenger(2023)}]{PCTNet}
Guerreiro, J. J.~A.; Nakazawa, M.; and Stenger, B. 2023.
\newblock {PCT}-{N}et: Full Resolution Image Harmonization Using Pixel-Wise
  Color Transformations.
\newblock In \emph{CVPR}.

\bibitem[{Guo et~al.(2022)Guo, Gu, Zheng, Dong, and Zheng}]{guo2022transformer}
Guo, Z.; Gu, Z.; Zheng, B.; Dong, J.; and Zheng, H. 2022.
\newblock Transformer for Image Harmonization and Beyond.
\newblock \emph{IEEE Transactions on Pattern Analysis and Machine
  Intelligence}.

\bibitem[{Guo et~al.(2021{\natexlab{a}})Guo, Guo, Zheng, Gu, Zheng, and
  Dong}]{guo2021image}
Guo, Z.; Guo, D.; Zheng, H.; Gu, Z.; Zheng, B.; and Dong, J.
  2021{\natexlab{a}}.
\newblock Image harmonization with transformer.
\newblock In \emph{ICCV}.

\bibitem[{Guo et~al.(2021{\natexlab{b}})Guo, Zheng, Jiang, Gu, and
  Zheng}]{guo2021intrinsic}
Guo, Z.; Zheng, H.; Jiang, Y.; Gu, Z.; and Zheng, B. 2021{\natexlab{b}}.
\newblock Intrinsic image harmonization.
\newblock In \emph{CVPR}.

\bibitem[{Hachnochi et~al.(2023)Hachnochi, Zhao, Orzech, Gal, Mahdavi-Amiri,
  Cohen-Or, and Bermano}]{cdc}
Hachnochi, R.; Zhao, M.; Orzech, N.; Gal, R.; Mahdavi-Amiri, A.; Cohen-Or, D.;
  and Bermano, A.~H. 2023.
\newblock Cross-domain Compositing with Pretrained Diffusion Models.
\newblock \emph{arXiv preprint arXiv:2302.10167}.

\bibitem[{Hang et~al.(2022)Hang, Xia, Yang, and Liao}]{hang2022scs}
Hang, Y.; Xia, B.; Yang, W.; and Liao, Q. 2022.
\newblock SCS-Co: Self-Consistent Style Contrastive Learning for Image
  Harmonization.
\newblock In \emph{CVPR}.

\bibitem[{Hao, Iizuka, and Fukui(2020)}]{Hao2020bmcv}
Hao, G.; Iizuka, S.; and Fukui, K. 2020.
\newblock Image Harmonization with Attention-based Deep Feature Modulation.
\newblock In \emph{BMVC}.

\bibitem[{He et~al.(2016)He, Zhang, Ren, and Sun}]{he2016deep}
He, K.; Zhang, X.; Ren, S.; and Sun, J. 2016.
\newblock Deep residual learning for image recognition.
\newblock In \emph{CVPR}.

\bibitem[{Huang et~al.(2023)Huang, An, Wei, Luo, and Pfister}]{quantart}
Huang, S.; An, J.; Wei, D.; Luo, J.; and Pfister, H. 2023.
\newblock QuantArt: Quantizing Image Style Transfer Towards High Visual
  Fidelity.
\newblock In \emph{CVPR}.

\bibitem[{Huang and Belongie(2017)}]{huang2017arbitrary}
Huang, X.; and Belongie, S. 2017.
\newblock Arbitrary style transfer in real-time with adaptive instance
  normalization.
\newblock In \emph{ICCV}.

\bibitem[{Huo et~al.(2021)Huo, Jin, Li, Wu, Lai, Shi, and
  Gao}]{huo2021manifold}
Huo, J.; Jin, S.; Li, W.; Wu, J.; Lai, Y.-K.; Shi, Y.; and Gao, Y. 2021.
\newblock Manifold alignment for semantically aligned style transfer.
\newblock In \emph{ICCV}.

\bibitem[{Jiang et~al.(2021)Jiang, Zhang, Zhang, Wang, Lin, Sunkavalli, Chen,
  Amirghodsi, Kong, and Wang}]{Jiang_2021_ICCV}
Jiang, Y.; Zhang, H.; Zhang, J.; Wang, Y.; Lin, Z.; Sunkavalli, K.; Chen, S.;
  Amirghodsi, S.; Kong, S.; and Wang, Z. 2021.
\newblock SSH: A Self-Supervised Framework for Image Harmonization.
\newblock In \emph{ICCV}.

\bibitem[{Jing et~al.(2020)Jing, Liu, Ding, Wang, Ding, Song, and
  Wen}]{jing2020dynamic}
Jing, Y.; Liu, X.; Ding, Y.; Wang, X.; Ding, E.; Song, M.; and Wen, S. 2020.
\newblock Dynamic instance normalization for arbitrary style transfer.
\newblock In \emph{AAAI}.

\bibitem[{Ke et~al.(2022)Ke, Sun, Zhu, Xu, and Lau}]{ke2022harmonizer}
Ke, Z.; Sun, C.; Zhu, L.; Xu, K.; and Lau, R.~W. 2022.
\newblock Harmonizer: Learning to Perform White-Box Image and Video
  Harmonization.
\newblock In \emph{ECCV}.

\bibitem[{Kolkin, Salavon, and Shakhnarovich(2019)}]{kolkin2019style}
Kolkin, N.; Salavon, J.; and Shakhnarovich, G. 2019.
\newblock Style transfer by relaxed optimal transport and self-similarity.
\newblock In \emph{CVPR}.

\bibitem[{Lai et~al.(2016)Lai, Huang, Hu, Ahuja, and Yang}]{lai2016comparative}
Lai, W.-S.; Huang, J.-B.; Hu, Z.; Ahuja, N.; and Yang, M.-H. 2016.
\newblock A comparative study for single image blind deblurring.
\newblock In \emph{CVPR}.

\bibitem[{Li and Wand(2016)}]{li2016combining}
Li, C.; and Wand, M. 2016.
\newblock Combining markov random fields and convolutional neural networks for
  image synthesis.
\newblock In \emph{CVPR}.

\bibitem[{Li et~al.(2018)Li, Liu, Kautz, and Yang}]{li2018learning}
Li, X.; Liu, S.; Kautz, J.; and Yang, M.-H. 2018.
\newblock Learning linear transformations for fast arbitrary style transfer.
\newblock \emph{arXiv preprint arXiv:1808.04537}.

\bibitem[{Li et~al.(2017)Li, Fang, Yang, Wang, Lu, and Yang}]{li2017universal}
Li, Y.; Fang, C.; Yang, J.; Wang, Z.; Lu, X.; and Yang, M.-H. 2017.
\newblock Universal style transfer via feature transforms.
\newblock \emph{NeurIPS}.

\bibitem[{Liang, Cun, and Pun(2022)}]{liang2021spatial}
Liang, J.; Cun, X.; and Pun, C.-M. 2022.
\newblock Spatial-Separated Curve Rendering Network for Efficient and
  High-Resolution Image Harmonization.
\newblock In \emph{ECCV}.

\bibitem[{Lin et~al.(2014)Lin, Maire, Belongie, Hays, Perona, Ramanan,
  Doll{\'a}r, and Zitnick}]{lin2014microsoft}
Lin, T.-Y.; Maire, M.; Belongie, S.; Hays, J.; Perona, P.; Ramanan, D.;
  Doll{\'a}r, P.; and Zitnick, C.~L. 2014.
\newblock Microsoft coco: Common objects in context.
\newblock In \emph{ECCV}.

\bibitem[{Ling et~al.(2021)Ling, Xue, Song, Xie, and Gu}]{ling2021region}
Ling, J.; Xue, H.; Song, L.; Xie, R.; and Gu, X. 2021.
\newblock Region-aware adaptive instance normalization for image harmonization.
\newblock In \emph{CVPR}.

\bibitem[{Liu et~al.(2023)Liu, Huynh, Chen, Arap, and Hamid}]{LEMaRT}
Liu, S.; Huynh, C.~P.; Chen, C.; Arap, M.; and Hamid, R. 2023.
\newblock LEMaRT: Label-Efficient Masked Region Transform for Image
  Harmonization.
\newblock In \emph{CVPR}.

\bibitem[{Liu et~al.(2021)Liu, Lin, He, Li, Wang, Li, Sun, Li, and
  Ding}]{liu2021adaattn}
Liu, S.; Lin, T.; He, D.; Li, F.; Wang, M.; Li, X.; Sun, Z.; Li, Q.; and Ding,
  E. 2021.
\newblock Adaattn: Revisit attention mechanism in arbitrary neural style
  transfer.
\newblock In \emph{ICCV}.

\bibitem[{Luan et~al.(2018)Luan, Paris, Shechtman, and Bala}]{luan2018deep}
Luan, F.; Paris, S.; Shechtman, E.; and Bala, K. 2018.
\newblock Deep painterly harmonization.
\newblock In \emph{CGF}.

\bibitem[{Meng et~al.(2021)Meng, He, Song, Song, Wu, Zhu, and Ermon}]{sdedit}
Meng, C.; He, Y.; Song, Y.; Song, J.; Wu, J.; Zhu, J.-Y.; and Ermon, S. 2021.
\newblock Sdedit: Guided image synthesis and editing with stochastic
  differential equations.
\newblock In \emph{ICLR}.

\bibitem[{Nichol(2016)}]{nichol2016painter}
Nichol, K. 2016.
\newblock Painter by numbers.
\newblock \url{https://www.kaggle.com/competitions/painter-by-numbers/data}.

\bibitem[{Park and Lee(2019)}]{park2019arbitrary}
Park, D.~Y.; and Lee, K.~H. 2019.
\newblock Arbitrary style transfer with style-attentional networks.
\newblock In \emph{CVPR}.

\bibitem[{Peng, Wang, and Wang(2019)}]{peng2019element}
Peng, H.-J.; Wang, C.-M.; and Wang, Y.-C.~F. 2019.
\newblock Element-Embedded Style Transfer Networks for Style Harmonization.
\newblock In \emph{BMVC}.

\bibitem[{Peng et~al.(2022)Peng, Luo, Liu, Zhang, Wang, Wang, Tai, Wang, and
  Lin}]{peng2022frih}
Peng, J.; Luo, Z.; Liu, L.; Zhang, B.; Wang, T.; Wang, Y.; Tai, Y.; Wang, C.;
  and Lin, W. 2022.
\newblock FRIH: Fine-grained Region-aware Image Harmonization.
\newblock \emph{arXiv preprint arXiv:2205.06448}.

\bibitem[{Ronneberger, Fischer, and Brox(2015)}]{ronneberger2015u}
Ronneberger, O.; Fischer, P.; and Brox, T. 2015.
\newblock U-net: Convolutional networks for biomedical image segmentation.
\newblock In \emph{MICCAI}.

\bibitem[{Sanakoyeu et~al.(2018)Sanakoyeu, Kotovenko, Lang, and
  Ommer}]{sanakoyeu2018style}
Sanakoyeu, A.; Kotovenko, D.; Lang, S.; and Ommer, B. 2018.
\newblock A style-aware content loss for real-time hd style transfer.
\newblock In \emph{ECCV}.

\bibitem[{Sheng et~al.(2018)Sheng, Lin, Shao, and Wang}]{sheng2018avatar}
Sheng, L.; Lin, Z.; Shao, J.; and Wang, X. 2018.
\newblock Avatar-net: Multi-scale zero-shot style transfer by feature
  decoration.
\newblock In \emph{CVPR}.

\bibitem[{Simonyan and Zisserman(2015)}]{VGG19}
Simonyan, K.; and Zisserman, A. 2015.
\newblock Very deep convolutional networks for large-scale image recognition.
\newblock \emph{ICLR}.

\bibitem[{Sofiiuk, Popenova, and Konushin(2021)}]{sofiiuk2021foreground}
Sofiiuk, K.; Popenova, P.; and Konushin, A. 2021.
\newblock Foreground-aware semantic representations for image harmonization.
\newblock In \emph{WACV}.

\bibitem[{Touvron et~al.(2023)Touvron, Bojanowski, Caron, Cord, El{-}Nouby,
  Grave, Izacard, Joulin, Synnaeve, Verbeek, and
  J{\'{e}}gou}]{TouvronBCCEGIJSVJ23}
Touvron, H.; Bojanowski, P.; Caron, M.; Cord, M.; El{-}Nouby, A.; Grave, E.;
  Izacard, G.; Joulin, A.; Synnaeve, G.; Verbeek, J.; and J{\'{e}}gou, H. 2023.
\newblock ResMLP: Feedforward Networks for Image Classification With
  Data-Efficient Training.
\newblock \emph{{IEEE} Trans. Pattern Anal. Mach. Intell.}, 45(4): 5314--5321.

\bibitem[{Tsai et~al.(2017)Tsai, Shen, Lin, Sunkavalli, Lu, and
  Yang}]{tsai2017deep}
Tsai, Y.-H.; Shen, X.; Lin, Z.; Sunkavalli, K.; Lu, X.; and Yang, M.-H. 2017.
\newblock Deep image harmonization.
\newblock In \emph{CVPR}.

\bibitem[{Valanarasu et~al.(2023)Valanarasu, Zhang, Zhang, Wang, Lin,
  Echevarria, Ma, Wei, Sunkavalli, and Patel}]{valanarasu2022interactive}
Valanarasu, J. M.~J.; Zhang, H.; Zhang, J.; Wang, Y.; Lin, Z.; Echevarria, J.;
  Ma, Y.; Wei, Z.; Sunkavalli, K.; and Patel, V.~M. 2023.
\newblock Interactive portrait harmonization.
\newblock \emph{ICLR}.

\bibitem[{Wang et~al.(2020)Wang, Li, Wang, Hu, and
  Yang}]{wang2020collaborative}
Wang, H.; Li, Y.; Wang, Y.; Hu, H.; and Yang, M.-H. 2020.
\newblock Collaborative distillation for ultra-resolution universal style
  transfer.
\newblock In \emph{CVPR}.

\bibitem[{Wang et~al.(2023)Wang, Gharbi, Zhang, Xia, and
  Shechtman}]{WangCVPR2023}
Wang, K.; Gharbi, M.; Zhang, H.; Xia, Z.; and Shechtman, E. 2023.
\newblock Semi-supervised Parametric Real-world Image Harmonization.
\newblock In \emph{CVPR}.

\bibitem[{Wu et~al.(2019)Wu, Kirillov, Massa, Lo, and
  Girshick}]{wu2019detectron2}
Wu, Y.; Kirillov, A.; Massa, F.; Lo, W.-Y.; and Girshick, R. 2019.
\newblock Detectron2.
\newblock \url{https://github.com/facebookresearch/detectron2}.

\bibitem[{Xing et~al.(2022)Xing, Li, Wang, Zhu, and Chen}]{xing2022composite}
Xing, Y.; Li, Y.; Wang, X.; Zhu, Y.; and Chen, Q. 2022.
\newblock Composite photograph harmonization with complete background cues.
\newblock In \emph{ACM MM}.

\bibitem[{Xue et~al.(2022)Xue, Ran, Chen, Jia, Zhao, and Tang}]{xue2022dccf}
Xue, B.; Ran, S.; Chen, Q.; Jia, R.; Zhao, B.; and Tang, X. 2022.
\newblock DCCF: Deep Comprehensible Color Filter Learning Framework for
  High-Resolution Image Harmonization.
\newblock In \emph{ECCV}.

\bibitem[{Yan et~al.(2022)Yan, Lu, Shuai, and Zhang}]{yan2022style}
Yan, X.; Lu, Y.; Shuai, J.; and Zhang, S. 2022.
\newblock Style Image Harmonization via Global-Local Style Mutual Guided.
\newblock In \emph{ACCV}.

\bibitem[{Zhang, Wen, and Shi(2020)}]{zhang2020deep}
Zhang, L.; Wen, T.; and Shi, J. 2020.
\newblock Deep image blending.
\newblock In \emph{WACV}.

\bibitem[{Zhang et~al.(2019)Zhang, Fang, Wang, Wang, Lin, Fu, and
  Yang}]{zhang2019multimodal}
Zhang, Y.; Fang, C.; Wang, Y.; Wang, Z.; Lin, Z.; Fu, Y.; and Yang, J. 2019.
\newblock Multimodal style transfer via graph cuts.
\newblock In \emph{ICCV}.

\bibitem[{Zhang et~al.(2023)Zhang, Huang, Tang, Huang, Ma, Dong, and Xu}]{inst}
Zhang, Y.; Huang, N.; Tang, F.; Huang, H.; Ma, C.; Dong, W.; and Xu, C. 2023.
\newblock Inversion-Based Creativity Transfer with Diffusion Models.
\newblock In \emph{CVPR}.

\bibitem[{Zhu et~al.(2022)Zhu, Zhang, Lin, Wu, Chai, and Guo}]{zhu2022image}
Zhu, Z.; Zhang, Z.; Lin, Z.; Wu, R.; Chai, Z.; and Guo, C.-L. 2022.
\newblock Image Harmonization by Matching Regional References.
\newblock \emph{arXiv preprint arXiv:2204.04715}.

\end{thebibliography}


\begin{thebibliography}{17}
\providecommand{\natexlab}[1]{#1}

\bibitem[{Cao, Hong, and Niu(2023)}]{cao2022painterly}
Cao, J.; Hong, Y.; and Niu, L. 2023.
\newblock Painterly Image Harmonization in Dual Domains.
\newblock \emph{AAAI}.

\bibitem[{Deng et~al.(2022)Deng, Tang, Dong, Ma, Pan, Wang, and
  Xu}]{deng2022stytr2}
Deng, Y.; Tang, F.; Dong, W.; Ma, C.; Pan, X.; Wang, L.; and Xu, C. 2022.
\newblock StyTr2: Image Style Transfer with Transformers.
\newblock In \emph{CVPR}.

\bibitem[{Goodfellow et~al.(2020)Goodfellow, Pouget-Abadie, Mirza, Xu,
  Warde-Farley, Ozair, Courville, and Bengio}]{goodfellow2020generative}
Goodfellow, I.; Pouget-Abadie, J.; Mirza, M.; Xu, B.; Warde-Farley, D.; Ozair,
  S.; Courville, A.; and Bengio, Y. 2020.
\newblock Generative adversarial networks.
\newblock \emph{Communications of the ACM}, 63(11): 139--144.

\bibitem[{Hachnochi et~al.(2023)Hachnochi, Zhao, Orzech, Gal, Mahdavi-Amiri,
  Cohen-Or, and Bermano}]{cdc}
Hachnochi, R.; Zhao, M.; Orzech, N.; Gal, R.; Mahdavi-Amiri, A.; Cohen-Or, D.;
  and Bermano, A.~H. 2023.
\newblock Cross-domain Compositing with Pretrained Diffusion Models.
\newblock \emph{arXiv preprint arXiv:2302.10167}.

\bibitem[{He et~al.(2016)He, Zhang, Ren, and Sun}]{he2016deep}
He, K.; Zhang, X.; Ren, S.; and Sun, J. 2016.
\newblock Deep residual learning for image recognition.
\newblock In \emph{CVPR}.

\bibitem[{Huang et~al.(2023)Huang, An, Wei, Luo, and Pfister}]{quantart}
Huang, S.; An, J.; Wei, D.; Luo, J.; and Pfister, H. 2023.
\newblock QuantArt: Quantizing Image Style Transfer Towards High Visual
  Fidelity.
\newblock In \emph{CVPR}.

\bibitem[{Kuznetsova et~al.(2020)Kuznetsova, Rom, Alldrin, Uijlings, Krasin,
  Pont-Tuset, Kamali, Popov, Malloci, Kolesnikov, Duerig, and
  Ferrari}]{Kuznetsova2020TheOI}
Kuznetsova, A.; Rom, H.; Alldrin, N.~G.; Uijlings, J. R.~R.; Krasin, I.;
  Pont-Tuset, J.; Kamali, S.; Popov, S.; Malloci, M.; Kolesnikov, A.; Duerig,
  T.; and Ferrari, V. 2020.
\newblock The Open Images Dataset V4.
\newblock \emph{International Journal of Computer Vision}, 128: 1956--1981.

\bibitem[{Lin et~al.(2014)Lin, Maire, Belongie, Hays, Perona, Ramanan,
  Doll{\'a}r, and Zitnick}]{lin2014microsoft}
Lin, T.-Y.; Maire, M.; Belongie, S.; Hays, J.; Perona, P.; Ramanan, D.;
  Doll{\'a}r, P.; and Zitnick, C.~L. 2014.
\newblock Microsoft coco: Common objects in context.
\newblock In \emph{ECCV}.

\bibitem[{Liu et~al.(2021)Liu, Lin, He, Li, Wang, Li, Sun, Li, and
  Ding}]{liu2021adaattn}
Liu, S.; Lin, T.; He, D.; Li, F.; Wang, M.; Li, X.; Sun, Z.; Li, Q.; and Ding,
  E. 2021.
\newblock Adaattn: Revisit attention mechanism in arbitrary neural style
  transfer.
\newblock In \emph{ICCV}.

\bibitem[{Luan et~al.(2018)Luan, Paris, Shechtman, and Bala}]{luan2018deep}
Luan, F.; Paris, S.; Shechtman, E.; and Bala, K. 2018.
\newblock Deep painterly harmonization.
\newblock In \emph{CGF}.

\bibitem[{Meng et~al.(2021)Meng, He, Song, Song, Wu, Zhu, and Ermon}]{sdedit}
Meng, C.; He, Y.; Song, Y.; Song, J.; Wu, J.; Zhu, J.-Y.; and Ermon, S. 2021.
\newblock Sdedit: Guided image synthesis and editing with stochastic
  differential equations.
\newblock In \emph{ICLR}.

\bibitem[{Nichol(2016)}]{nichol2016painter}
Nichol, K. 2016.
\newblock Painter by numbers.
\newblock \url{https://www.kaggle.com/competitions/painter-by-numbers/data}.

\bibitem[{Park and Lee(2019)}]{park2019arbitrary}
Park, D.~Y.; and Lee, K.~H. 2019.
\newblock Arbitrary style transfer with style-attentional networks.
\newblock In \emph{CVPR}.

\bibitem[{Peng, Wang, and Wang(2019)}]{peng2019element}
Peng, H.-J.; Wang, C.-M.; and Wang, Y.-C.~F. 2019.
\newblock Element-Embedded Style Transfer Networks for Style Harmonization.
\newblock In \emph{BMVC}.

\bibitem[{Simonyan and Zisserman(2015)}]{VGG19}
Simonyan, K.; and Zisserman, A. 2015.
\newblock Very deep convolutional networks for large-scale image recognition.
\newblock \emph{ICLR}.

\bibitem[{Wu et~al.(2019)Wu, Kirillov, Massa, Lo, and
  Girshick}]{wu2019detectron2}
Wu, Y.; Kirillov, A.; Massa, F.; Lo, W.-Y.; and Girshick, R. 2019.
\newblock Detectron2.
\newblock \url{https://github.com/facebookresearch/detectron2}.

\bibitem[{Zhang et~al.(2023)Zhang, Huang, Tang, Huang, Ma, Dong, and Xu}]{inst}
Zhang, Y.; Huang, N.; Tang, F.; Huang, H.; Ma, C.; Dong, W.; and Xu, C. 2023.
\newblock Inversion-Based Creativity Transfer with Diffusion Models.
\newblock In \emph{CVPR}.

\end{thebibliography}

\end{document}

\maketitle

In this document, we provide additional materials to support our main paper. In Section~\ref{sec:training_set_preparation}, we describe more details for our training data preparation. In Section~\ref{sec:ablation}, we conduct ablation studies and provide visualization results for the ablated versions. In Section~\ref{sec:out_of_COCO}, we demonstrate the generalization ability of our method.  In Section~\ref{sec:cmp_with_baseline}, we show more visualization results compared with baseline methods.  In Section~\ref{sec:failure_case}, we discuss the failure case of our method.

\section{More Details of Training Data} \label{sec:training_set_preparation}

Following previous works~\cite{peng2019element,cao2022painterly}, we use COCO \cite{lin2014microsoft} and WikiArt \cite{nichol2016painter}.
COCO \cite{lin2014microsoft} contains instance segmentation annotations for 80 object categories, while WikiArt \cite{nichol2016painter} contains digital artistic paintings from different styles.
We create composite images based on the training sets of these two datasets, with the photographic objects from COCO and the painterly backgrounds from WikiArt.

As introduced in Section 3.1 in the main paper, we use off-the-shelf  object detection model \cite{wu2019detectron2} pretrained on COCO \cite{lin2014microsoft} to detect $34,570$ objects from the artistic paintings in the training set of WikiArt. 
For each painterly object, we aim to retrieve the photographic objects with similar appearance and semantics automatically. 

\begin{figure}[t]
\centering
\includegraphics[width=0.8\linewidth]{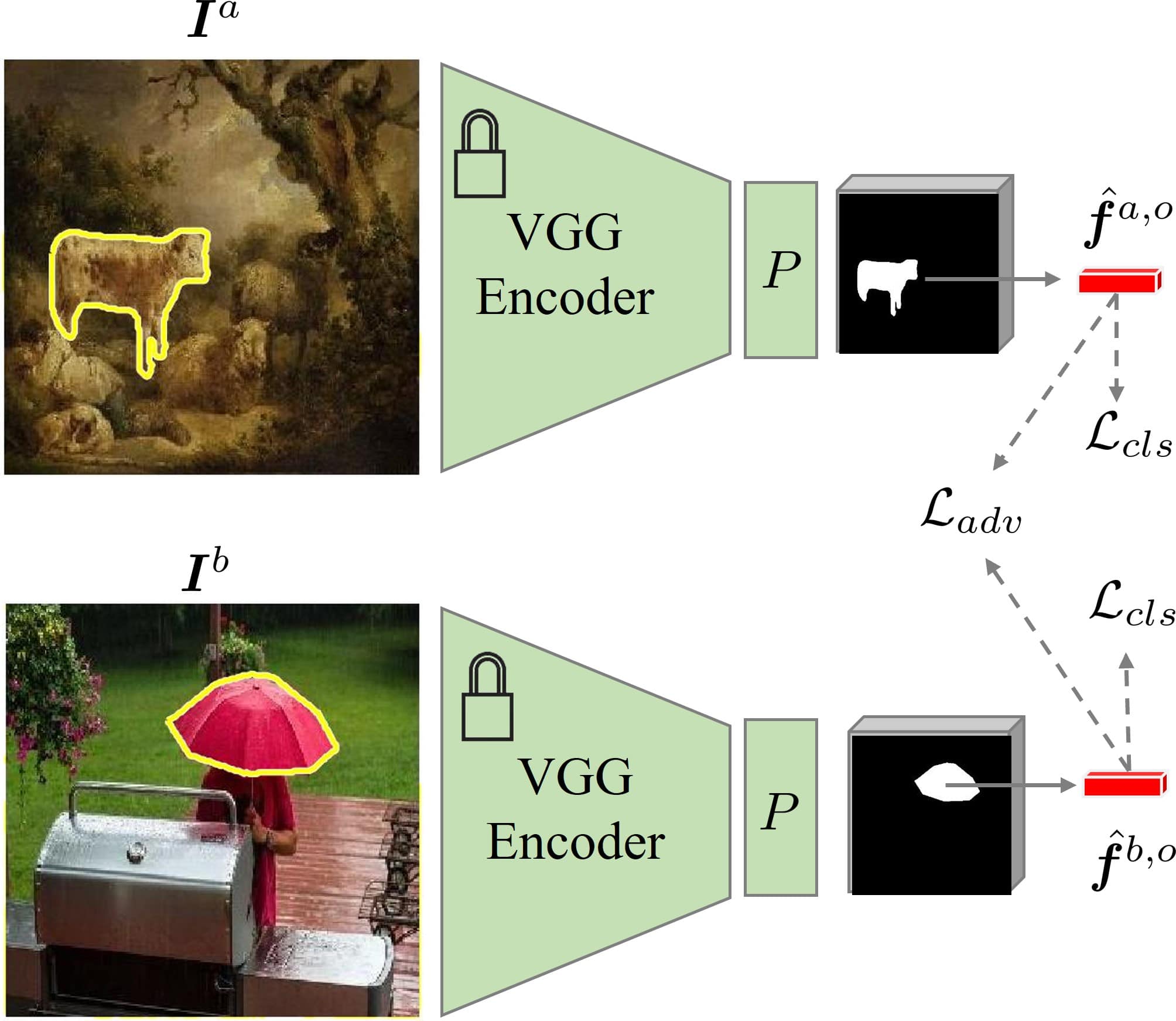}
\caption{The network structure for object retrieval. We use VGG-19 \cite{VGG19} encoder and projection module $P$ to extract object feature $\hat{\bm{f}}^{a,o}$ (\emph{resp.}, $\hat{\bm{f}}^{b,o}$) from  painterly (\emph{resp.}, photographic) object in $\bm{I}^a$ (\emph{resp.}, $\bm{I}^b$). }
\label{fig:object_retrieval}
\end{figure}

\begin{figure*}[t]
\centering
\includegraphics[width=0.99\linewidth]{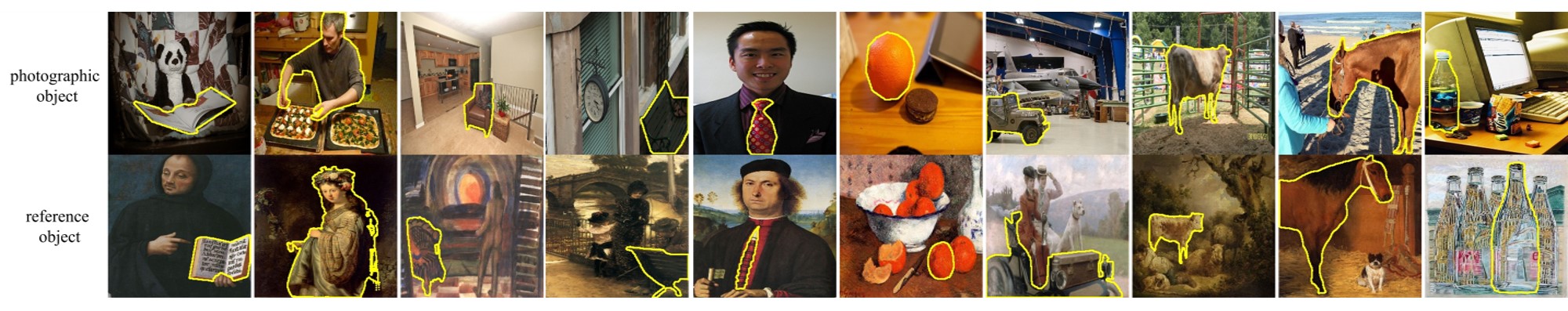}
\caption{Example annotated pairs of photographic objects and reference objects, in which the objects are outlined in yellow. Paired objects in each column have similar color and semantics.}
\label{fig:reference_object_supp}
\end{figure*}

\begin{figure*}[t]
\centering
\includegraphics[width=0.95\linewidth]{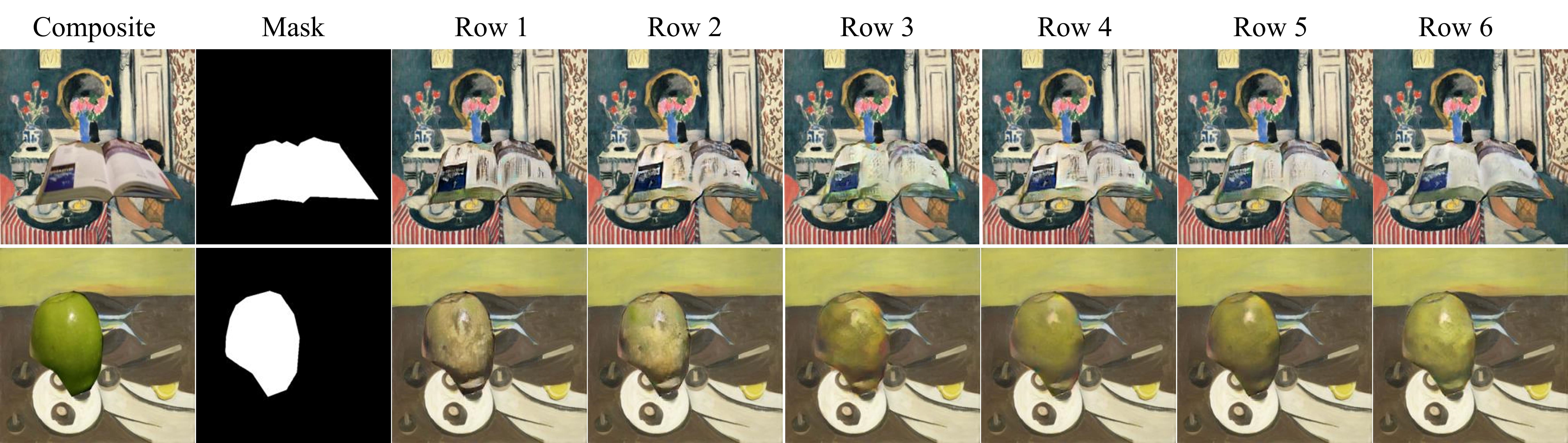}
\caption{From left to right, we show the composite image, composite foreground mask, the harmonized results of row 1-6 in Table \ref{tab:ablation}.}
\label{fig:ablation_studies}
\end{figure*}

To achieve this goal, we design a simple object retrieval network as illustrated in Figure~\ref{fig:object_retrieval}, which is similar to our main network. Specifically, we use pretrained VGG-19 network~\cite{VGG19} followed by projection module $P$ (a residual block \cite{he2016deep}) to produce feature maps for both artistic paintings and photographic images, in which $P$ aims to project two domains into a common domain. Given an artistic painting $\bm{I}^a$ and a photographic image $\bm{I}^b$, we extract their feature maps and perform average pooling within the foreground region, producing the painterly object feature $\hat{\bm{f}}^{a,o}$ and the photographic object feature $\hat{\bm{f}}^{b,o}$. To pull close the object features from two domains, we employ adversarial loss $\mathcal{L}_{adv}$ \cite{goodfellow2020generative} to make the painterly object features indistinguishable from photographic object features. Moreover, to preserve the discriminative information of objects, we also apply classification loss $\mathcal{L}_{cls}$ to object features. For photographic objects in COCO dataset \cite{lin2014microsoft}, we have the ground-truth category labels provided by \cite{lin2014microsoft}. For painterly objects, we use the category labels predicted by the object detection model \cite{wu2019detectron2}. For photographic object features and painterly object features, we use the same classifier so that the object features from the same category yet different domains can be grouped together. 
Thus, the total loss to train the object retrieval network can be written as $\mathcal{L}_{ret}=\mathcal{L}_{adv}+\mathcal{L}_{cls}$.

After training the object retrieval network, for each painterly object, we retrieve $100$ nearest photographic objects based on $L_2$ distance between their object features. Nevertheless, the retrieved results are very noisy and far from usable. Therefore, we only treat the retrieved objects as candidate objects, and ask human annotators to filter out dissimilar candidate objects. After filtering, we have $33,294$ painterly objects associated with similar photographic objects, and
each painterly object has an average of $9.83$ similar photographic objects. Given a pair of painterly object and its similar photographic object, we refer to the painterly object as the reference object of this photographic object.
In Figure \ref{fig:reference_object_supp}, we show several example photographic objects and their reference objects. It can be seen that the objects cover a wide range of categories including person, animal, vehicle, furniture, and so on. The photographic objects have similar color and semantics with their reference objects.

\begin{table}[t] 
\centering
\begin{tabular}{c|c|c}
\hline
Row & Method  & B-T score \\
\hline
1 &  w/o ObAdaIN &  -0.475\\
2 & w/o $\hat{\bm{f}}^{co}$/$\hat{\bm{f}}^{po}$  & -0.289\\
3 & w/o $\mathcal{L}_{obj}$  & 0.0278\\
4 & w/o $\mathcal{L}_{map}^p$   & 0.162 \\
5 & w/o $\mathcal{L}_{map}^c$  & 0.211 \\
6 & Full  & 0.363 \\
\hline
\end{tabular}
\caption{Results of different ablated versions of our ArtoPIH. }
\label{tab:ablation}
\end{table}

\begin{figure}[t]
\centering
\includegraphics[width=0.8\linewidth]{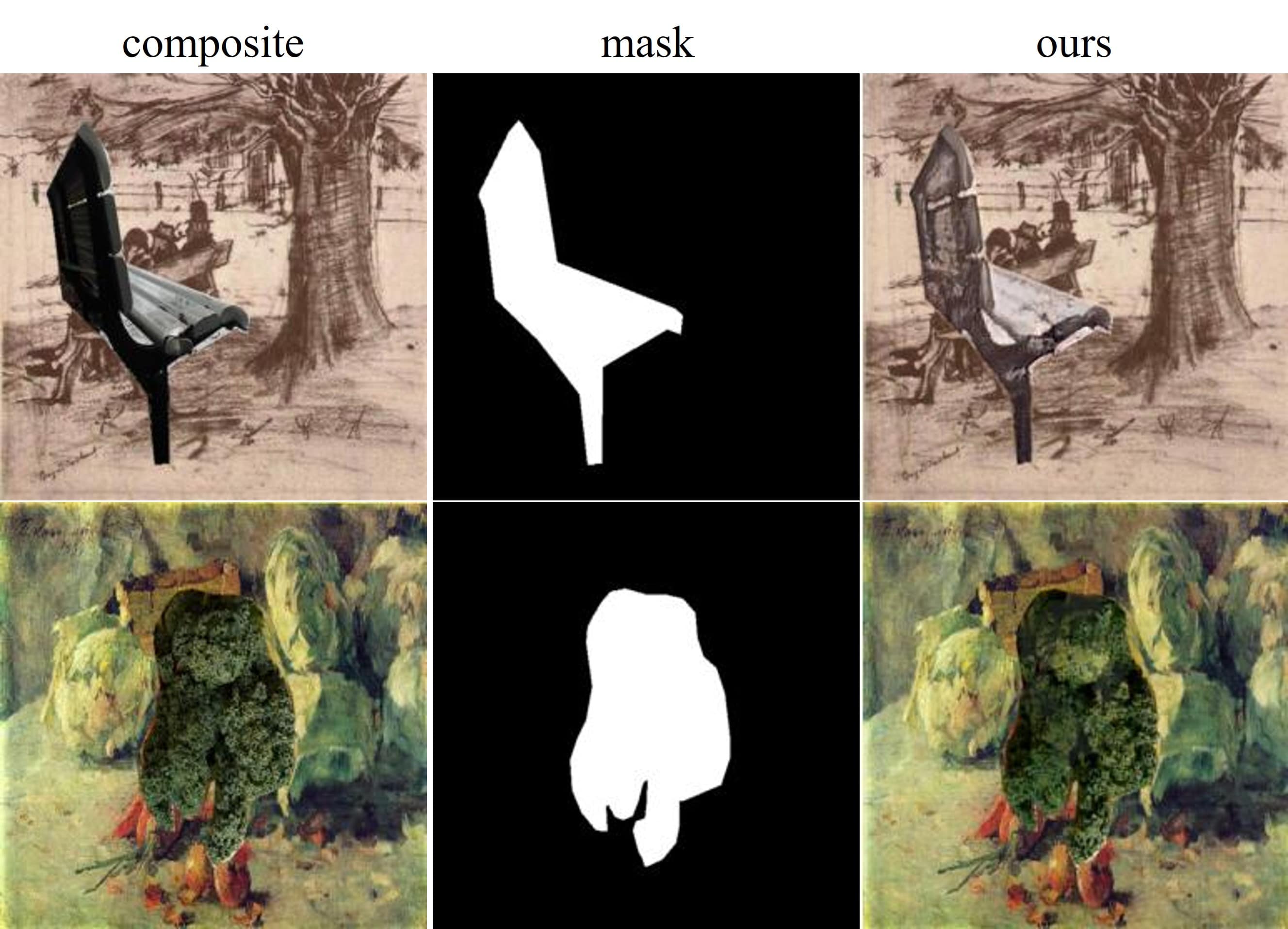}
\caption{Two example failure cases of our ArtoPIH.}
\label{fig:failure_cases}
\end{figure}

\begin{figure}[t]
\centering
\includegraphics[width=0.92\linewidth]{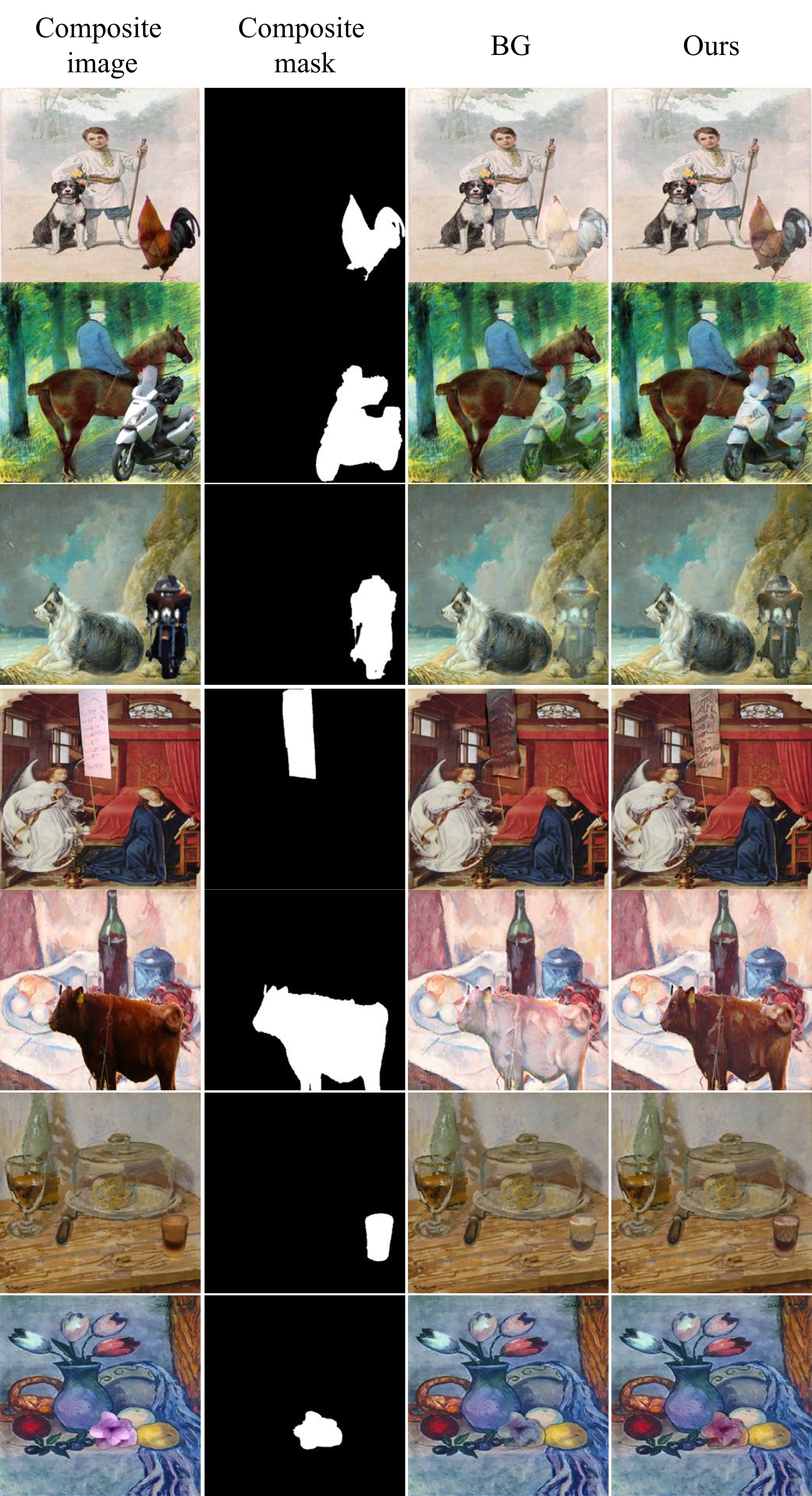}
\caption{From left to right, we show the composite image, composite foreground mask, the harmonized result using background style vector, and our harmonized result.}
\label{fig:out_of_COCO}
\end{figure}

\begin{figure*}[h]
\centering
\includegraphics[width=0.92\linewidth]{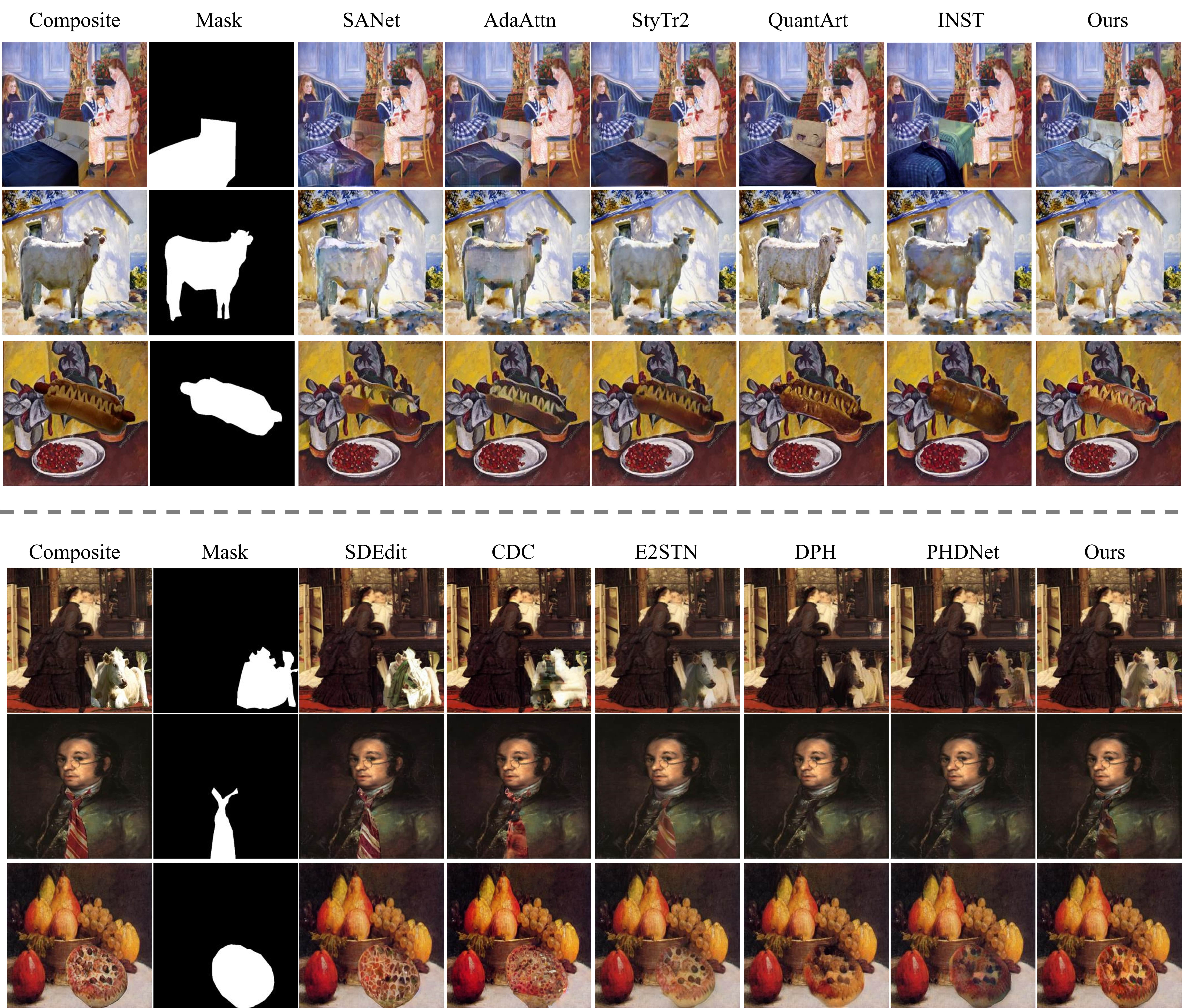}
\caption{In the upper part, we compare with style transfer baselines SANet~\cite{park2019arbitrary}, AdaAttN~\cite{liu2021adaattn}, StyTr2~\cite{deng2022stytr2}, QuantArt~\cite{quantart}, INST~\cite{inst}. 
In the lower part, we compare with painterly image harmonization baselines SDEdit~\cite{sdedit}, CDC~\cite{cdc}, E2STN~\cite{peng2019element}, DPH~\cite{luan2018deep}, PHDNet~\cite{cao2022painterly}.
 }
\label{fig:baseline_supp}
\end{figure*}

\section{Ablation Studies} \label{sec:ablation}

We first build a basic model and then gradually add our designed modules. First, we replace our ObAdaIN module with vanilla AdaIN, by applying background style to the composite object, leading to the basic model in row 1. 
Then, we learn the mapping from background style to object style without using object feature $\hat{\bm{f}}^{co}$/$\hat{\bm{f}}^{po}$, leading to row 2. Next, we add object features in the mapping, leading to our full method in row 6. 
Next, we ablate each loss term $\mathcal{L}_{obj}$, $\mathcal{L}_{map}^p$, and $\mathcal{L}_{map}^c$ in row 3, 4, and 5, respectively. 

We report the user study results (B-T score) for different ablated versions in Table \ref{tab:ablation}. We observe that row 1 achieves the worst performance by directly using background style. Row 2 is only slightly better than row 1 because the object style is not conditioned on the specific object information.  By comparing row 3-5 with row 6, we see that each ablated loss term causes performance drop, which proves that three loss terms contribute to the final performance altogether.

Next, we show the visualization results of different ablated versions in Figure \ref{fig:ablation_studies}. It can be seen that the row 1 and row 2 could only produce very poor results with notable artifacts,  because they directly transfer background style or hallucinate unsuitable target styles. Row 3-5 are better than row 1-2, but worse than row 6, which proves that each loss term can help improve the performance.

\section{Results beyond COCO Dataset} \label{sec:out_of_COCO}
To demonstrate the generalization ability of our ArtoPIH, we collect  photographic objects from Open Images \cite{Kuznetsova2020TheOI} dataset, which are out of $80$ COCO \cite{lin2014microsoft} object categories. Then, we create composite images using the collected photographic objects and artistic paintings from WikiArt \cite{nichol2016painter} test set. We show the harmonized results of our method in Figure~\ref{fig:out_of_COCO}. For comparison, we also show the harmonized results obtained using background style vector (BG), in the same way as Figure 5 in the main paper. Note that some exhibited categories may have similar ones in COCO dataset. For example, ``beer" in row 6 is similar to ``cup" in COCO, and ``whiteboard" in row 4 is similar to ``book" in COCO. Some other categories (\emph{e.g.}, ``rooster" in row 1) are farther from their similar ones in COCO. For all these categories, our method can produce more faithful and visually pleasant images than ``BG", which verifies that our ArtoPIH can generalize well to the foreground objects out of COCO dataset. 

\section{More Visual Comparison with Baselines} \label{sec:cmp_with_baseline}

In Section 4.2 in the main paper, we compare with two groups of baselines: artistic style transfer baselines and painterly image harmonization baselines.  

We show the comparison with two groups of baselines in Figure \ref{fig:baseline_supp}. Compared with the first group of baselines, we observe that our method can usually produce more visually pleasing objects which seem to appear naturally on the background. The harmonized results of baselines are prone to  suffer from unreasonable color, noticeable artifacts, or under-stylization. Compared with the second group of baselines, our method can produce well-harmonized foreground object while preserving the original object information. In contrast, the harmonized results from baselines are likely to have unsatisfactory stylization effect, or destroy the original object information.

\section{Failure Cases} \label{sec:failure_case}

Although our ArtoPIH can usually achieve satisfactory results, there also exist some failure cases. For example, as shown in Figure~\ref{fig:failure_cases}, when the background has unified color style, the harmonized foreground is likely to have color mismatch with the background, because the predicted style vector may deviate from the background style vector in terms of color information. One possible solution is first detecting whether the background has unified color style. If so, we can directly apply background style vector to the foreground to be on the safe side.

\section*{Acknowledgments}
The work was supported by the National Natural Science Foundation of China (Grant No. 62076162), the Shanghai Municipal Science and Technology Major/Key Project, China (Grant No. 2021SHZDZX0102, Grant No. 20511100300).

\bibliography{supp.bbl}